\documentclass[twoside,11pt]{article}
\pdfoutput=0
\usepackage{standalone}
\usepackage{bm}
\usepackage{amsmath}
\usepackage{amssymb}
\usepackage{mathtools}
\usepackage{graphicx}
\usepackage{jmlr2e}
\usepackage{xifthen}
\usepackage{pstricks}
\usepackage{pst-math}
\usepackage{pst-func}
\usepackage{pst-plot}
\usepackage{snapshot}
\mathtoolsset{showonlyrefs}

\jmlrheading{14}{2013}{1801-1835}{9/12; Revised 4/13}{7/13}{Matthew J.\ Urry and Peter Sollich}
\ShortHeadings{Gaussian Processes on Random Graphs}{Urry and Sollich}
\firstpageno{1801}

\def\Sref #1{Section \ref{#1}}
\def\rmd {\textrm{d}}
\def\rmi {\textrm{i}}

\def\CoVar{\textrm{Cov}}
\def\T{^\textrm{T}}
\def\f{\bm{f}}
\def\g{\bm{g}}
\def\C{\bm{C}}
\def\x{\bm{x}}
\def\y{\bm{y}}
\def\z{\bm{z}}
\def\M{\bm{M}}
\def\K{\bm{K}}
\def\I{\bm{I}}
\def\m{\bm{m}}

\def\e0{\bm{e}_0}
\newrgbcolor{darkgreen}{0 0.7 0}
\renewcommand{\ni}[1][]{\ifthenelse{\isempty{#1}}{\gamma}{\gamma_{#1}}}
\begin{document}
\pagebreak
\title{Random Walk Kernels and Learning Curves for Gaussian Process Regression on Random Graphs}
\author{\name Matthew J. Urry \email matthew.urry@kcl.ac.uk \\
    \name Peter Sollich \email peter.sollich@kcl.ac.uk \\
    \addr Department of Mathematics\\
    King's College London\\
    London, WC2R 2LS, U.K.
    }
    \editor{Manfred Opper}
\maketitle

\begin{abstract}%
We consider learning on graphs, guided by kernels that encode similarity between vertices. Our focus is on random walk kernels, the analogues of squared exponential kernels in Euclidean spaces. We show that on large, locally treelike graphs these have some counter-intuitive properties, specifically in the limit of large kernel lengthscales. We consider using these kernels as covariance functions of Gaussian processes. In this situation one typically scales the prior globally to normalise the average of the prior variance across vertices. We demonstrate that,
in contrast to the Euclidean case, this generically leads to significant variation in the prior variance across vertices, which is undesirable from a probabilistic modelling point of view.
We suggest the random walk kernel should be normalised locally, so that each vertex has the same prior variance, and analyse the consequences of this by studying learning curves for Gaussian process regression. Numerical calculations as well as novel theoretical predictions for the learning curves using belief propagation show that one obtains distinctly different probabilistic models depending on the choice of normalisation. Our method for predicting the learning curves using belief propagation is significantly more accurate than previous approximations and should become exact in the limit of large random graphs.
\end{abstract}
\section{Introduction}\label{sec:intro}
\emph{Gaussian processes (GPs)} have become a workhorse for probabilistic inference that has been developed in a wide range of research fields under various guises \citep[see for example][]{Kleijnen2009, Handcock1993,Neal1996,Meinhold1983}. Their success and wide adoption can be attributed mainly to their intuitive nature and ease of use. They owe their intuitiveness to being one of a large family of kernel methods that implicitly map lower dimensional spaces with non-linear relationships to higher dimensional spaces where (hopefully) relationships are linear. This feat is achieved by using a kernel, which also encodes the types of functions that the GP prefers a priori. The ease of use of GPs is due to the simplicity of implementation, at least in the basic setting, where prior and posterior distributions are both Gaussian and can be written explicitly.

An important question for any machine learning method is how `quickly' the method can generalise its prediction of a rule to the entire domain of the rule (i.e., how many examples are required to achieve a particular generalisation error). This is encapsulated in the \emph{learning curve}, which traces average error versus number of examples. Learning curves have been studied for a variety of inference methods and are well understood for parametric models \citep{Seung1992,Amari1992,Watkin1993,Opper1995,Haussler1996,Freeman1997} but rather less is known for non-parametric models such as GPs. In the case of GP regression, research has predominantly focused on leaning curves for input data from Euclidean spaces \citep{Sollich1999a, Sollich1999b,
Opper1999, Williams2000, Malzahn2003, Sollich2002a,
Sollich2002b, Sollich2005}, but there are many domains for which the input data has a discrete structure. One of the simplest cases is the one where inputs are vertices on a graph, with connections on the graph encoding similarity relations between different inputs. Examples could include the internet, social networks, protein networks and financial markets. Such discrete input spaces with graph structure are becoming more important, and therefore so is an understanding of GPs, and machine learning techniques in general, on these spaces.

In this paper we expand on earlier work in \citet{Sollich2009} and \citet{Urry2010} and focus on predicting the learning curves of GPs used for regression (where outputs are from the whole real line) on large sparse graphs, using the \emph{random walk kernel} \citep{Kondor2002,Smola2003}.

The rest of this paper will be structured as follows. In \Sref{sec:randomwalk} we begin by analysing the random walk kernel, in particular with regard to the dependence on its lengthscale parameter, and study the approach to the fully correlated limit. With a better understanding of the random walk kernel in hand, we proceed in \Sref{sec:lc} to an analysis of the use of the random walk kernel for GP regression on graphs. We begin in \Sref{sec:kernelnorm} by looking at how kernel normalisation affects the prior probability over functions. We show that the more frequently used global normalisation of a kernel by its average prior variance is inappropriate for the highly location dependent random walk kernel, and suggest normalisation to uniform local prior variance as a remedy. To understand how this affects GP regression using random walk kernels quantitatively, we extend first in \Sref{sec:evalpred} an existing approximation to the learning curve in terms of kernel eigenvalues \citep{Sollich1999a,Opper2002} to the discrete input case, allowing for arbitrary normalisation. This approximation turns out to be accurate only in the initial and asymptotic regimes of the learning curve.

The core of our analysis begins in \Sref{sec:cavitypred} with the development of an improved learning curve approximation based on belief propagation. We first apply this, in \Sref{sec:cavityglobal}, to the case of globally normalised kernels as originally proposed. The belief propagation analysis for global normalisation also acts as a useful warm-up for the extension to the prediction of learning curves for the locally normalised kernel setting, which we present in \Sref{sec:cavitylocal}. In both sections we compare our predictions to numerical simulations, finding good agreement that improves significantly on the eigenvalue approximation. Finally, to emphasise the distinction between the use of globally and locally normalised kernels in GP regression, we study qualitatively the case of model mismatch, with a GP with a globally normalised kernel as the teacher and a GP with a locally normalised kernel as the student, or visa versa. The resulting learning curves show that the priors arising from the two different normalisations are fundamentally different; the learning curve can become non-monotonic and develop a maximum as a result of the mismatch. We conclude in \Sref{sec:conclusions} by summarising our results and discussing further potentially fruitful avenues of research.

\subsection{Main Results}
In this paper we will derive three key results; that normalisation of a kernel by its average prior variance leads to a complicated relationship between the prior variances and the local graph structure; that by fixing the scale to be equal everywhere using a local prescription $C_{ij} = \hat{C}_{ij}/\sqrt{\hat{C}_{ii}\hat{C}_{jj}}$ results in a fundamentally different probabilistic model; and that we can derive accurate predictions of the learning curves of Gaussian processes on graphs with a random walk kernel for both normalisations over a broad range of graphs and parameters. The last result is surprising since in continuous spaces this is only possible for a few very restrictive cases.

\section{The Random Walk Kernel} \label{sec:randomwalk}

A wide range of machine learning techniques like Gaussian processes capture prior correlations between points in an input space by mapping to a higher dimensional space, where correlations can be represented by a linear combination of `features' \citep[see, e.g.,][]{Rasmussen2005,Muller2001,Cristianini2000}. Direct calculation of correlations in this high dimensional space is avoided using the `kernel trick', where the kernel function implicitly calculates inner products in feature space. The widespread use of, and therefore extensive research in, kernel based machine learning has resulted in kernels being developed for a wide range of input spaces \citep[see][and references therein]{Genton2002}. We focus in this paper on the class of kernels introduced in \citet{Kondor2002}. These make use of the normalised graph Laplacian to define correlations between vertices of a graph.

We denote a generic graph by $G(\mathcal{V},\mathcal{E})$ with a vertex set $\mathcal{V}=\{1,\ldots,V\}$ and edge set $\mathcal{E}$. We encode the connection structure of $G$ using an adjacency matrix $\bm{A}$, where $A_{ij}=1$ if vertex $i$ is connected to $j$, and $0$ otherwise; we exclude self-loops so that $A_{ii}=0$. We denote the number of edges connected to vertex $i$, known as the degree, by $d_i=\sum_j A_{ij}$ and define the degree matrix $\bm{D}$ as a diagonal matrix of the vertex degrees, that is, $D_{ij}=d_i \delta_{ij}$. The class of kernels created in \citet{Kondor2002} is constructed using the \emph{normalised Laplacian}, $\bm{L} = \bm{I} - \bm{D}^{-1/2}\bm{A}\bm{D}^{-1/2}$ \citep[see][]{Chung1996} as a replacement for the Laplacian in continuous spaces. Of particular interest is the diffusion kernel and its easier to calculate approximation, the random walk kernel. Both of these kernels can be viewed as an approximation to the ubiquitous squared exponential kernel that is used in continuous spaces. The direct graph equivalent of the squared exponential kernel is given by the \emph{diffusion kernel} \citep{Kondor2002}. It is defined as
\begin{equation}\label{eqn:diffkernel}
\bm{C} = \exp\left(-\frac{1}{2}\sigma^{2}\bm{L}\right),\quad \sigma > 0,
\end{equation}
where $\sigma$ sets the length-scale of the kernel. Unlike in continuous spaces, the exponential in the diffusion kernel is costly to calculate. To avoid this, \citet{Smola2003} proposed as a cheaper approximation the \emph{random walk kernel}
\begin{equation}\label{eqn:randomwalkkernel}
\bm{C} = \left(\bm{I} - a^{-1}\bm{L}\right)^{p} = \left( (1-a^{-1})\bm{I} +
a^{-1}\bm{D}^{-1/2}\bm{A}\bm{D}^{-1/2}\right)^{p},\quad a>2,\quad p\in\mathbb{N}.
\end{equation}
This gives back the diffusion kernel in the limit $a,p\to\infty$ whilst keeping $p/a=\sigma^{2}/2$ fixed. The random walk kernel derives its name from its use of random walks to express correlations between vertices. Explicitly, a binomial expansion of Equation~\eqref{eqn:randomwalkkernel} gives
\begin{equation}\label{eqn:binomrandomwalk}
\begin{split}
\bm{C} &= \sum_{q=0}^{p}\binom{p}{q}(1-a^{-1})^{p-q}(a^{-1})^q(\bm{D}^{-1/2}\bm{A}\bm{D}^{-1/2})^{q} \\&=
\bm{D}^{-1/2}\sum_{q=0}^{p}\binom{p}{q}(1-a^{-1})^{p-q}(a^{-1})^q(\bm{A}\bm{D}^{-1})^{q}\bm{D}^{1/2}.
\end{split}
\end{equation}
The matrix $\bm{A}\bm{D}^{-1}$ is a random walk transition matrix: $(\bm{A}\bm{D}^{-1})_{ij}$ is the probability of being at vertex $i$ after one random walk step starting from vertex $j$. Apart from the pre- and post-multiplication by $\bm{D}^{-1/2}$ and $\bm{D}^{1/2}$, the kernel $\bm{C}$ is therefore a $q$-step random walk transition matrix, averaged over the number of steps $q$ distributed as
$q\sim\textrm{Binomial}(p,a^{-1})$. Equivalently one can interpret the random walk kernel as a $p$-step lazy random walk, where at each step the walker stays at the current vertex with probability $(1-a^{-1})$ and moves to a neighbouring vertex with probability $a^{-1}$.

Using either interpretation, one sees that $p/a$ is the lengthscale over which the random walk can diffuse along the graph, and hence the lengthscale describing the typical maximum range of the random walk kernel. In the limit of large $p$, where this lengthscale diverges, the kernel should represent full correlation across all vertices. One can see that this is the case by observing that for large $p$, a random walk on a graph will approach its stationary distribution, $\bm{p}_{\infty}\propto \bm{D}\bm{e}$, $\bm{e}=(1,\dots,1)^{T}$. The $q$-step transition matrix for large $q$ is therefore $(\bm{A}\bm{D}^{-1})^q \approx \bm{p}_{\infty}\bm{e}^{T} = \bm{D}\bm{e}\bm{e}^{T}$, representing the fact that the random walk becomes stationary independently of the starting vertex. This gives, for $p\to\infty$, the kernel $\bm{C}\propto \bm{D}^{1/2}\bm{e}\bm{e}^{T}\bm{D}^{1/2}$, that is, $C_{ij} \propto d_i^{1/2}d_{j}^{1/2}$. This corresponds to full correlation across vertices as expected; explicitly, if $\bm{f}$ is a Gaussian process on the graph with covariance matrix  $\bm{D}^{1/2}\bm{e}\bm{e}^{T}\bm{D}^{1/2}$, then  $\bm{f}=v\bm{D}^{1/2}\bm{e}$ with $v$ a single Gaussian degree of freedom.

We next consider how random walk kernels on graphs approach the fully correlated case, and show that even for `simple' graphs the convergence to this limit is non-trivial. Before we do so, we note an additional peculiarity of random walk kernels compared to their Euclidean counterparts: in addition to the maximum range lengthscale $p/a$ discussed so far, they have a diffusive lengthscale $\sigma=(2p/a)^{1/2}$, which is suggested for large $p$ and $a$ by the lengthscale of the corresponding diffusion kernel \eqref{eqn:diffkernel}. This diffusive lengthscale will appear in our analysis of learning curves in the large $p$-limit \Sref{sec:scaling}.

\subsection{The $d$-Regular Tree: A Concrete Example} \label{sec:dregtree}
To begin our discussion of the dependence of the random walk kernel on the lengthscale $p/a$, we first look at how this kernel behaves on a $d$-regular graph sampled uniformly from the set of all $d$-regular graphs. Here $d$-regular means that all vertices have degree $d_i=d$.
For a large enough number of vertices $V$, typical cycles in such a $d$-regular graph are also large, of length $O(\log V)$, and can be neglected for calculation of the kernel when $V\to\infty$. We therefore begin by assuming the graph is an infinite tree, and assess later how the cycles that do exist on random $d$-regular graphs cause departures from this picture.

A $d$-regular tree is a graph where each vertex has degree $d$ with \emph{no cycles}; it is unique up to permutations of the vertices. Since all vertices on the tree are equivalent, the random walk kernel $C_{ij}$ can only depend on the distance between vertices $i$ and $j$, that is, the smallest number of steps on the graph required to get from one vertex to the other. Denoting the value of a $p$-step lazy random walk kernel for vertices a distance $l$ apart by $C_{l,p}$, we can determine these values by recursion over $p$ as follows:
\begin{equation}\label{eqn:shelljump}
\begin{split}
C_{l,p=0} &= \delta_{l,0}, \qquad \gamma_{p+1}C_{0,p+1} = \left(1-\frac{1}{a}\right)C_{0,p} + \frac{1}{a}C_{1,p},\\
\gamma_{p+1}C_{l,p+1} &= \frac{1}{ad}C_{l-1,p} + \left(1-\frac{1}{a}\right)C_{l,p} + \frac{d-1}{ad}C_{l+1,p}\quad l\geq1.
\end{split}
\end{equation}
Here $\gamma_{p}$ is chosen to achieve the desired normalisation of the prior variance for every $p$. We will normalise so that $C_{0,p}=1$.

\begin{figure}
\input{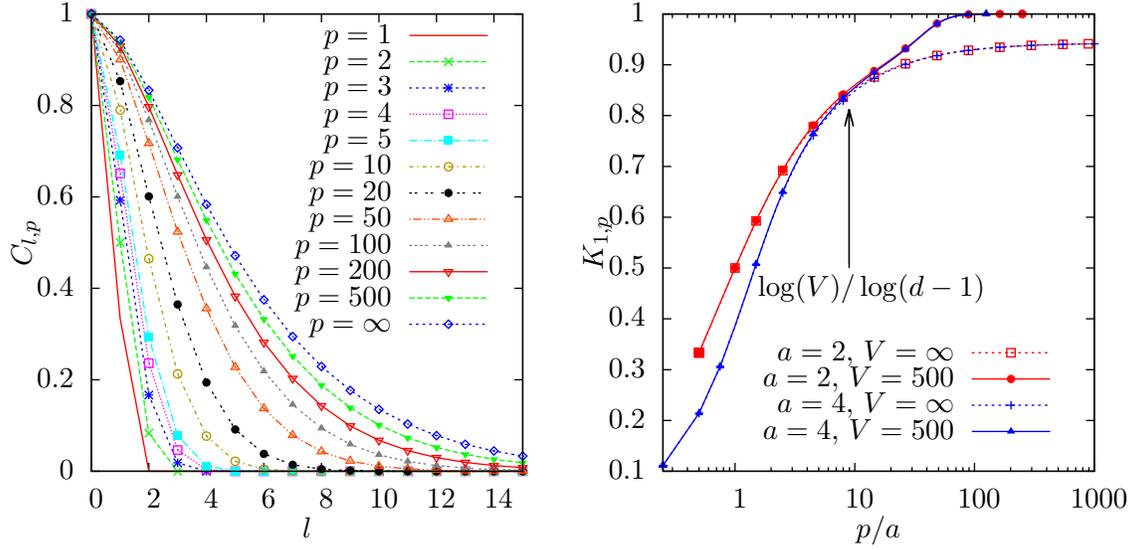}
\caption{(Left) Random walk kernel $C_{l,p}$ on a 3-regular tree plotted against distance $l$ for increasing number of steps $p$ and $a=2$. (Right) Comparison between numerical results for the average nearest neighbour kernel $K_{1,p}$ on random 3-regular graphs with the result $C_{1,p}$ on a 3-regular tree, calculated numerically by iteration of \protect\eqref{eqn:shelljump}.
}\label{fig:clp}\label{fig:treegraph}
\end{figure}

Figure \ref{fig:clp} (left) shows the results obtained by iterating Equation~\eqref{eqn:shelljump} numerically for a 3-regular tree with $a=2$. As expected the kernel becomes longer-ranged initially as $p$ is increased, but seems to approach a non-trivial limiting form. This can be calculated analytically and is given by (see Appendix \ref{app:clplc})
\begin{equation}\label{eqn:clpinfinity}
C_{l,p\to\infty} = \left[ 1+\frac{l(d-2)}{d}\right]\frac{1}{(d-1)^{l/2}}.
\end{equation}
Equation \eqref{eqn:clpinfinity} can be derived by taking the $\sigma^2\to\infty$ limit of the integral expression for the diffusion kernel from \citet{Chung1999} whilst preserving normalisation of the kernel (see Appendix \ref{app:chunglimit} for further details). Alternatively the result (\ref{eqn:clpinfinity}) can be obtained by rewriting the random walk in terms of shells, that is, grouping vertices according to distance $l$ from a chosen central vertex. The number of vertices in the $l$-th shell, or shell volume, is $v_l = d(d-1)^{l-1}$ for $l\geq 1$ and $v_0=1$. Introducing $R_{l,p} = C_{l,p}\sqrt{v_l}$, Equation~\eqref{eqn:shelljump} can be written in the form
\begin{equation}\label{eqn:Rshelljump}
\begin{split}
R_{l,p=0} &= \delta_{l,0}, \qquad \gamma_{p+1}R_{0,p+1} = \left(1-\frac{1}{a}\right)R_{0,p} + \frac{1}{a\sqrt{d}}R_{1,p},\\
\gamma_{p+1}R_{l,p+1} &= \frac{\sqrt{d-1}}{ad}R_{l-1,p} + \left(1-\frac{1}{a}\right)R_{l,p} + \frac{\sqrt{d-1}}{ad}R_{l+1,p}\quad l\geq1.
\end{split}
\end{equation}
This is just the un-normalised diffusion equation for a biased random walk on a one dimensional lattice with a reflective boundary at 0. This has been solved in \citet{Monthus1996}, and mapping this solution back to $C_{l,p}$ gives \eqref{eqn:clpinfinity} (see Appendix \ref{app:lcscale} for further details).

To summarise thus far, the analysis on a $d$-regular tree shows that, for large $p$, the random walk kernel does not approach the expected fully correlated limit: because all vertices have the same degree this limit would correspond to $C_{l,p\to\infty}=1$. On the other hand, on a $d$-regular graph with any finite number $V$ of vertices, the fully correlated limit must necessarily be approached as $p\to\infty$. As a large regular graph is locally treelike, the difference must arise from the existence of long cycles in a regular graph.

To estimate when the existence of cycles will start to affect the kernel, consider first a $d$-regular tree truncated at depth $l$. This will have
$V=1+\sum_{i=1}^{l}d(d-1)^{i-1} = O(d(d-1)^{l-1})$ vertices. On a $d$-regular graph with the same number of vertices, we therefore expect to encounter cycles after a number of steps, taken along the graph, of order $l$. In the random walk kernel the typical number of steps is $p/a$, so effects of cycles should appear once $p/a$ becomes larger than
\begin{equation}\label{eqn:p_over_a_limit}
  \frac{p}{a} \approx \frac{\log(V)}{\log(d-1)}.
\end{equation}
Figure \ref{fig:treegraph} (right) shows a comparison between $C_{1,p}$ as calculated from Equation~\eqref{eqn:shelljump} for a $3$-regular tree and its analogue on random $3$-regular graphs of finite size, which we call $K_{1,p}$. We define this analogue as the average of $C_{ij}/\sqrt{C_{ii}C_{jj}}$ over all pairs of neighbouring vertices on a fixed graph, averaged further over a number of randomly generated regular graphs. The square root accounts for the fact that local kernel values $C_{ii}$ can vary slightly on a regular graph because of cycles, while they are the same for all vertices of a regular tree. Looking at Figure \ref{fig:treegraph} (right) one sees that, as expected from the arguments above, the nearest neighbour kernel value for the $3$-regular graph, $K_{1,p}$, coincides with its analogue $C_{1,p}$ on the $3$-regular tree for small $p$. When $p/a$ crosses the threshold \eqref{eqn:p_over_a_limit}, cycles in the regular graph become important and the two curves separate. For larger $p$, the kernel value for neighbouring vertices approaches the fully correlated limit $K_{1,p}\to 1$ on a regular graph, while on a regular tree one has the non-trivial limit $C_{1,p}\to 2\sqrt{d-1}/d$ from \eqref{eqn:clpinfinity}.

In conclusion of our analysis of random walk kernels, we have seen that these kernels have an unusual dependence on their lengthscale $p/a$. In particular, kernel values for vertices a short distance apart can remain significantly below the fully correlated limit, even if $p/a$ is large. That limit is approached only once $p/a$ becomes larger than the graph size-dependent threshold \eqref{eqn:p_over_a_limit}, at which point cycles become important. We have focused here on random regular graphs, but the same qualitative behaviour should be observed also on graphs with a non-trivial distribution of vertex degrees $d_i$.

\section{Learning Curves for Gaussian Process Regression}\label{sec:lc}
Having reached a better understanding of the random walk kernel we now study its application in machine learning. In particular we focus on the use of the random walk kernel for regression with Gaussian processes. We will begin, for completeness, with an introduction to GPs for regression. For a more comprehensive discussion of GPs for machine learning we direct the reader to \citet{Rasmussen2005}.

\subsection{Gaussian Process Regression: Kernels as Covariance Functions}

Gaussian process regression is a Bayesian inference technique that constructs a posterior distribution over a function space, $P(f|\x,\y)$, given training input locations $\x=(x_{1},\ldots,x_{N})\T$ and corresponding function value outputs $\y=(y_{1},\ldots,y_{N})\T$. The posterior is constructed from a prior distribution $P(f)$ over the function space and the likelihood $P(\y|f,\x)$ to generate the observed output values from function $f$ by using Bayes' theorem
\begin{equation}
P(f|\x,\y) = \frac{P(\y|f,\x)P(f)}{\int \rmd f' P(\y|f',\x)P(f') }.
\end{equation}
In the GP setting the prior is chosen to be a Gaussian process, where any finite number of function values has a joint Gaussian distribution, with a covariance matrix with entries given by a \emph{covariance function} or kernel $C(x,x')$ and with a mean vector with entries given by a \emph{mean function} $\mu(x)$. For simplicity we will focus on zero mean GPs\footnote{In the discussion and analysis that follows, generalisation to non-zero mean GPs is straightforward.} and a Gaussian likelihood, which amounts to assuming that training outputs are corrupted by independent and identically distributed Gaussian noise. Under these assumptions all distributions are Gaussian and can be calculated explicitly. If we assume we are given training data $\{(x_{\mu},y_{\mu})|\mu=1,\ldots,N\}$ where $y_\mu$ is the value of the target or `teacher' function at input location $x_{\mu}$, corrupted by additive Gaussian noise with variance $\sigma^{2}$, the posterior distribution is then given by another Gaussian process with mean and covariance functions
\begin{align}
  \bar{f}(x) &=\bm{k}(x)\T\bm{K}^{-1}\bm{y},\label{eqn:GPmean}\\
  \CoVar(x,x') &=
C(x,x')-\bm{k}(x)\T\bm{K}^{-1}\bm{k}(x'),\label{eqn:GPcovariance}
\end{align}
where $\bm{k}(x) = (C(x_{1},x),\ldots,C(x_{N},x))\T$ and $K_{\mu\nu} = C(x_{\mu},x_{\nu})+ \delta_{\mu\nu}\sigma^{2}$. With the posterior in the form of a Gaussian process, predictions are simple. Assuming a squared loss function, the optimal prediction of the outputs is given by $\bar{f}(x)$ and a measure of uncertainty in the prediction is provided by $\CoVar(x,x)^{1/2}$.

Equations \eqref{eqn:GPmean} and \eqref{eqn:GPcovariance} illustrate that, in the setting of GP regression, kernels are used to change the type of function preferred by the Gaussian process prior, and correspondingly the posterior. The kernel can encode prior beliefs about smoothness properties, lengthscale and expected amplitude of the function we are trying to predict. Of particular importance for the discussion below, $C(x,x)$ gives the prior variance of the function $f$ at input $x$, so that $C(x,x)^{1/2}$ sets the typical function \emph{amplitude} or \emph{scale}.

\subsection{Kernel Normalisation}\label{sec:kernelnorm}
Conventionally one fixes the desired scale of the kernel using a \emph{global normalisation}: denoting the unnormalised kernel by $\hat{C}(x,x')$ one scales $C(x,x')= \hat{C}(x,x')/\kappa$ to achieve a desired average of $C(x,x)$ across input locations $x$. In Euclidean spaces one typically uses translationally invariant kernels like the squared exponential kernel. For these, $C(x,x)$ is the same for all input locations $x$ and so global normalisation is sufficient to fix a spatially uniform scale for the prior amplitude. In the case of kernels on graphs, on the other hand, the local connectivity structure around each vertex can be different. Since information about correlations `propagates' only along graph edges, graph kernels are not generally translation invariant. In particular, there can be large variation among the prior variances at different vertices. This is usually undesirable in a probabilistic model, unless one has strong prior knowledge to justify such variation. For the random walk kernel, the local prior variances are the diagonal entries of Equation~\eqref{eqn:binomrandomwalk}. These are directly related to the probability of return of a lazy random walk on a graph, which depends sensitively on the local graph structure. This dependence is in general non-trivial, and not just expressible through, for example, the degree of the local vertex. It seems difficult to imagine a scenario where such a link between prior variances and local graph structures could be justified by prior knowledge.

\begin{figure}
  \input{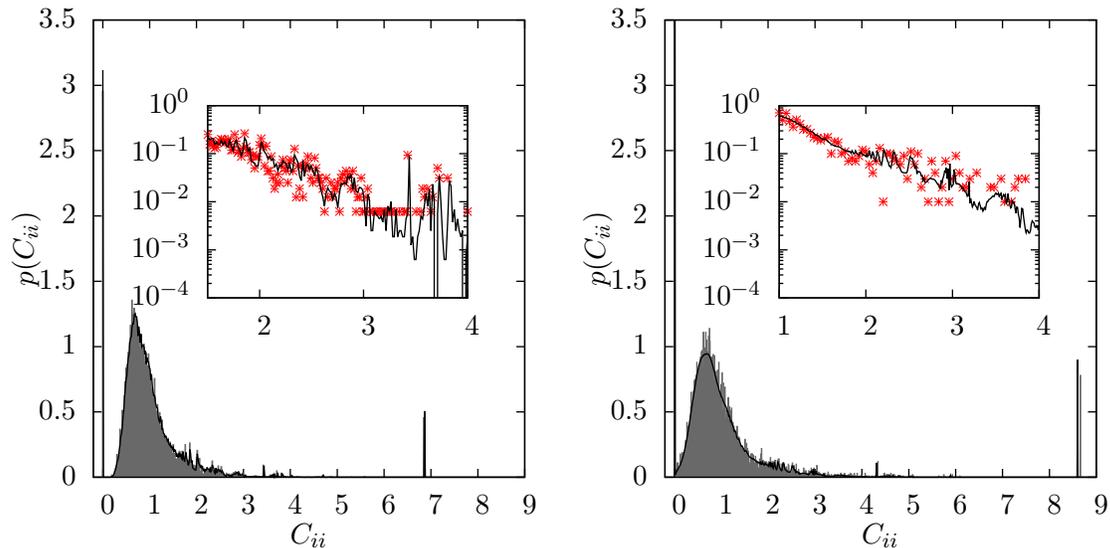}
\caption{(Left) Grey: histogram of prior variances for the globally normalised random walk kernel with $a=2$, $p=10$ on
a single instance of an Erd\H{o}s-R\'enyi graph with mean degree $\lambda=3$ and $V=10000$ vertices. Black: prediction
for this distribution in the large graph limit (see Section \protect\ref{sec:cavityvariance}). Inset: Linear-log plot of the
tail of the distribution.  (Right) As (left) but for a power law generalised random graph with exponent 2.5 and cutoff 2.\label{fig:poissonvariance}\label{fig:powerlawvariance}\label{fig:priorvariance}}
\end{figure}

To emphasise the issue, Figure \ref{fig:poissonvariance} shows examples of distributions of local prior variances $C_{ii}$ for random walk kernels globally normalised to an average prior variance of unity.\footnote{We use $C_{ii}$ again here, instead of $C(i,i)$ as in our general discussion of GPs; the subscript notation is more intuitive because the covariance function on a graph is just a $V\times V$ matrix.} The distributions are peaked around the desired value of unity but contain many `outliers' from vertices with abnormally low or high prior variance. Figure \ref{fig:poissonvariance} (left) shows the distribution of $C_{ii}$ for a large single instance of an Erd\H{o}s-R\'enyi random graph \citep{Erdos1959}. In such graphs, each edge is present independently of all others with some fixed probability, giving a Poisson distribution of degrees $p_{\lambda}(d) = \lambda^{d}\exp(-\lambda)/d!$; for the figure we chose average degree $\lambda=3$. Figure \ref{fig:powerlawvariance} (right) shows analogous results for a generalised random graph with power law mixing distribution \citep{Britton2006}. Generalised random graphs are an extension of Erd\H{o}s-R\'enyi random graphs where different edges are assigned different probabilities of being present. By appropriate choice of these probabilities \citep{Britton2006}, one can generate a degree distribution that is a superposition of Poisson distributions, $p(d)=\int \rmd \lambda\, p_{\lambda}(d)p(\lambda)$. We have taken a shifted Pareto distribution, $p(\lambda)=\alpha\lambda^{\alpha}_{m}/\lambda^{\alpha+1}$ with exponent $\alpha=2.5$ and lower cutoff $\lambda_m=2$ for the distribution of the means.

Looking first at Figure \ref{fig:poissonvariance} (left), we know that large Erd\H{o}s-R\'enyi graphs are locally tree-like and hence one might expect that this would lead to relatively uniform local prior variances. As shown in the figure, however, even for such tree-like graphs large variations can exist in the local prior variances. To give some specific examples, the large spike near 0 is caused by single disconnected vertices and the smaller spike at around 6.8 arises from two-vertex (single edge) disconnected subgraphs. Single vertex subgraphs have an atypically small prior variance since, for a single disconnected vertex $i$, before normalisation $C_{ii} = (1-a^{-1})^{p}$ which is the $q=0$ contribution from Equation~\eqref{eqn:binomrandomwalk}. Other vertices in the graph will get additional contributions from $q\geq 1$ and so have a larger prior variance. This effect will become more pronounced as $p$ is increased and the binomial weights assign less weight to the $q=0$ term.

Somewhat surprisingly at first sight, the opposite effect is seen for two-vertex disconnected subgraphs as shown by the spike around $C_{ii}=6.8$ in Figure \ref{fig:poissonvariance} (left). For vertices on such subgraphs, $C_{ii} = \sum_{q=0}^{\lfloor p/2\rfloor} \binom{p}{2q}a^{-2q}(1-a^{-1})^{p-2q}$, which is an atypically large return probability: after any even number of steps, the walker must always return to its starting vertex. A similar situation would occur on vertices at the centre of a star.  This illustrates that local properties of a vertex alone, like its degree, do not constrain the prior variance. In a two-vertex disconnected subgraph both vertices have degree 1. But there will generically be other vertices of degree 1 that are dangling ends of a large connected graph component, and these will not have similarly elevated return probabilities. Thus, local graph structure is intertwined in a complex manner with local prior variance.

The black line in Figure \ref{fig:poissonvariance} (left) shows theoretical predictions (see Section \ref{sec:cavityvariance}) for the prior variance distribution in the large graph limit. There is significant fine structure in the various peaks, on which theory and simulations agree well where the latter give reliable statistics. The decay from the mean is roughly exponential (see linear-log plot in inset), emphasizing that the distribution of local prior variances is not only rather broad but can also have large tails.

For the power law random graph, Figure \ref{fig:poissonvariance} (right), the broad features of the distribution of local prior variances $C_{ii}$ are similar: a peak at the desired value of unity, overlaid by spikes which again come from single and two-vertex disconnected subgraphs. The inset shows that the tail beyond the mean is roughly exponential again, but with a slower decay; this is to be expected since power law graphs exhibit many more different local structures with a significantly larger probability than is the case for Erd\H{o}s-R\'enyi graphs. Accordingly, the distribution of the $C_{ii}$ also has a larger standard deviation than for the Erd\H{o}s-R\'enyi case. The maximum values of $C_{ii}$ that we see in these two specific graph instances follow the same trend, with $\max_i C_{ii}\approx 40$ for the power law graph and $\max_i C_{ii}\approx 15$ for the Erd\H{o}s-R\'enyi graph. Such large values would constitute rather unrealistic prior assumptions about the scaling of the target function at these vertices.

To summarise, Figure \ref{fig:priorvariance} shows that after global normalisation a random walk kernel can retain a large spread in the local prior variances, with the latter depending on the graph structure in a complicated manner. We propose that to overcome this one should use a \emph{local normalisation}. For a desired prior variance $c$ this means normalising according to $C_{ij} = c\hat{C}_{ij}/(\kappa_{i}\kappa_{j})^{1/2}$ with local normalisation constants $\kappa_i = \hat{C}_{ii}$; here $\hat{C}_{ij}$ is the unnormalised kernel matrix as before.  This guarantees that all vertices have exactly equal prior variance as in the Euclidean case, that is, all vertices have a prior variance of $c$. No uncontrolled local variation in the scaling of the function prior then remains, and the computational overhead of local over global normalisation is negligible. Graphically, if we were to normalise the kernel to unity according to the local prescription, a plot of prior variances like the one in Figure \ref{fig:priorvariance} would be a delta peak centred at 1.

The effect of this normalisation on the behaviour of GP regression is a key question for the remainder of this paper; numerical simulation results are shown in \Sref{sec:predictinglc} below, while our theoretical analysis is described in \Sref{sec:cavitypred}.

\subsection{Predicting the Learning Curve}\label{sec:predictinglc}
The performance of non-parametric methods such as GPs can be characterised by studying the \emph{learning curve},
\begin{equation}
\epsilon(N) = \Bigg\langle \Bigg\langle\Bigg\langle\Bigg\langle\frac{1}{V}\sum_{i=1}^{V}\left(g_i - \langle
f_i\rangle_{\f|\x,\y}\right)^{2}
    \Bigg\rangle_{\y|\g,\x}\Bigg\rangle_{\g}\Bigg\rangle_{\x}\Bigg\rangle_{\mathcal{G}},
\end{equation}
defined as the average squared error between the student and teacher's predictions $\f = (f_1,\ldots,f_V)\T$ and $\g=(g_1,\ldots,g_V)\T$ respectively, averaged over the student's posterior distribution given the data $\f|\x,\y$, the outputs given the teacher $\y|\g,\x$, the teacher functions $\g$, and the input locations $\x$. This gives the average generalisation error as a function of the number of training examples. For simplicity we will assume that the input distribution is uniform across the vertices of the graph.

Because we are analysing GP regression on graphs, after the averages discussed so far the generalisation error will still depend on the structure of the specific graph considered. We therefore include an additional average, over all graphs in a random graph ensemble $\mathcal{G}$. We consider graph ensembles defined by the distribution of degrees $d_i$: we specify a degree sequence $\{d_1, \ldots, d_V\}$, or, for large $V$, equivalently a degree distribution $p(d)$, and pick uniformly at random any one of the graphs that has this degree distribution. The actual shape of the degree distribution is left arbitrary, as long as it has finite mean. Our analysis therefore has broad applicability, including in particular the graph types already mentioned above ($d$-regular graphs, where $p(d')=\delta_{dd'}$, Erd\H{o}s-R\'enyi graphs, power law generalised random graphs).

For this paper, as is typical for learning curve studies, we will assume that teacher and student have the same prior distribution over functions, and likewise that the assumed Gaussian noise of variance $\sigma^2$ reflects the actual noise process corrupting the training data. This is known as the \emph{matched case}.\footnote{The case of mismatch has been considered in \citet{Malzahn2005} for fixed teacher functions, and for prior and noise level mismatch in \citet{Sollich2002a,Sollich2005}.} Under this assumption the generalisation error becomes the Bayes error, which given that we are considering squared error simplifies to the posterior variance of the student averaged over data sets and graphs \citep{Rasmussen2005}. Since we only need the posterior variance we shift $\f$ so that the posterior mean is $\bm{0}$; $f_i$ is then just the deviation of the function value at vertex $i$ from the posterior mean. The Bayes error can now be written as
\begin{equation}\label{eqn:bayeserror}
  \epsilon(N) = \Bigg\langle\Bigg\langle\Bigg\langle
\frac{1}{V}\sum_{i=1}^{V}f_i^{2}\Bigg\rangle_{\f|\x}\Bigg\rangle_{\x}\Bigg\rangle_{\mathcal{G}}.
\end{equation}
Note that by shifting the posterior distribution to zero mean, we have eliminated the dependence on $\y$ in the above equation. That this should be so can also be seen from \eqref{eqn:GPcovariance} for the posterior (co-)variance, which only depends on training inputs $\x$ but not the corresponding outputs $\y$.

The averages in Equation \eqref{eqn:bayeserror} are in general difficult to calculate analytically, because the training input locations $\x$ enter in a highly nonlinear matter, see~\eqref{eqn:GPcovariance}; only for very specific situations can exact results be obtained \citep{Malzahn2005,Rasmussen2005}.
Approximate learning curve predictions have been derived, for Euclidean input spaces, with some degree of success \citep{Sollich1999a, Sollich1999b,
Opper1999, Williams2000, Malzahn2003, Sollich2002a,
Sollich2002b, Sollich2005}. We will show that in the case of GP regression for functions defined on graphs, learning curves can be predicted exactly in the limit of large random graphs. This prediction is broadly applicable because the degree distribution that specifies the graph ensemble is essentially arbitrary.

It is instructive to begin our analysis by extending a previous approximation seen in \citet{Sollich1999a} and \cite{Malzahn2005} to our discrete graph case. In so doing we will see explicitly how one may improve this approximation to fully exploit the structure of random graphs, using belief propagation or equivalently the \emph{cavity method} \citep{Mezard2003}. We will sketch the derivation of the existing approximation following the method of \citet{Malzahn2005}; the result given by \citet{Sollich1999a} is included in this as a somewhat more restricted approximation. Both the approximate treatment and our cavity method take a statistical mechanics approach, so we begin by rewriting Equation~\eqref{eqn:bayeserror} in terms of a \emph{generating} or \emph{partition function} $Z$
\begin{equation}\label{eqn:epgZ}
  \epsilon(N) = \left\langle\frac{1}{V}\sum_{i}\int \rmd\f P(\f|\x) f_i^{2}\right\rangle_{\x,\mathcal{G}} = -\lim_{\lambda\to0}\frac{2}{V}\frac{\partial}{\partial\lambda}\left\langle \log(Z)\right\rangle_{\x,\mathcal{G}},
\end{equation}
with
\begin{equation}
  Z = \int \rmd\f \exp\left(-\frac{1}{2}\f\T \C^{-1}\f - \frac{1}{2\sigma^{2}}\sum_{\mu=1}^Nf_{x_{\mu}}^2-\frac{\lambda}{2}\sum_{i}f_{i}^{2}\right).
\end{equation}
In this representation the inputs $\x$ only enter $Z$ through the sum over $\mu$. We introduce $\ni[i]$ to count the number of examples at vertex $i$ so that $Z$ becomes
\begin{equation}\label{eqn:Z}
  Z = \int \rmd\f \exp\left(-\frac{1}{2}\f\T\C^{-1}\f - \frac{1}{2}\f\T\textrm{diag}\left(\frac{\ni[i]}{\sigma^{2}}+\lambda\right)\f\right).
\end{equation}
The average in Equation~\eqref{eqn:epgZ} of the logarithm of this partition function can still not be carried out in closed form. The approximation given by \citet{Malzahn2005} and our present cavity approach diverge
at this point. Section \ref{sec:evalpred} discusses the existing approximation for the learning curve, applied to the case of regression on
a graph. Section \ref{sec:cavitypred} then improves on this using the cavity method to fully exploit the graph structure.

\subsection{Kernel Eigenvalue Approximation}\label{sec:evalpred}

The approach of \citet{Malzahn2005} is to average $\log(Z)$ from \eqref{eqn:Z} using the replica trick \citep{Mezard1987}.
One writes $\langle\log Z\rangle_{\x} = \lim_{n\to0}\frac{1}{n}\log\langle Z^{n}\rangle_{\x}$, performing the average $\langle Z^{n}\rangle_{\x}$ for integer $n$ and assuming that a continuation to $n\to0$ is possible. The required $n$-th power of Equation~\eqref{eqn:Z} is given by
\begin{equation}
\langle Z^n\rangle_{\x} = \int \prod_{a=1}^n
\rmd\f^a\left\langle\exp\left(-\frac{1}{2}\sum_a (\f^a)\T\C^{-1}\f^a
-\frac{1}{2\sigma^{2}}\sum_{i,a}\ni[i] (f_{i}^a)^{2}
-\frac{\lambda}{2}\sum_{i,a}(f_{i}^{a})^{2}\right)\right\rangle_{\x},
\end{equation}
where the replica index $a$ runs from $1$ to $n$.
Assuming as before that examples are generated independently and uniformly from $\mathcal{V}$, the data set average over $\x$ will, for large $V$, become equivalent to independent Poisson averages over $\ni[i]$ with mean $\nu=N/V$. Explicitly performing these averages gives
\begin{equation}
\langle Z^n\rangle_{\x} = \int \prod_{a=1}^n
\rmd\f^a\exp\left(-\frac{1}{2}\sum_a (\f^a)\T\C^{-1}\f^a
+\nu \sum_i \left(e^{-\sum_{a}(f_{i}^a)^{2}/2\sigma^{2}}-1\right)
-\frac{\lambda}{2}\sum_{i,a}(f_{i}^{a})^{2}\right).
\label{eqn:Opper2}
\end{equation}
In order to evaluate \eqref{eqn:Opper2} one has to find a way to deal with the exponential term in the exponent. \citet{Malzahn2005} do this using a variational approximation for the distribution of the $\bm{f}^a$, of Gaussian form. Eventually this leads to the following eigenvalue learning curve approximation (see also \citealp{Sollich1999a}):
\begin{equation}\label{eqn:mercerapprox}
  \epsilon(N) = g\left(\frac{N}{\epsilon(N)+\sigma^{2}}\right),\qquad g(h)=\sum_{\alpha=1}^{V}\left(\lambda_{\alpha}^{-1}+h\right)^{-1}.
\end{equation}
The eigenvalues $\lambda_\alpha$ of the kernel are defined here from the eigenvalue equation\footnote{Here and below we consider the case of a uniform distribution of inputs across vertices, though the results can be generalised to the non-uniform case.} $(1/V) \sum_j C_{ij} \phi_j = \lambda \phi_i$. The Gaussian variational approach is evidently justified for large $\sigma^2$, where a Taylor expansion of the exponential term in \eqref{eqn:Opper2} can be truncated after the quadratic term. For small noise levels, on the other hand, the Gaussian variational approach will in general not capture all the details of the fluctuations in the numbers of examples $\ni[i]$. This issue is expected to be most prominent for values of $\nu$ of order unity, where fluctuations in the number of examples are most relevant because some vertices will not have seen examples locally or nearby and will have posterior variance close to the prior variance, whereas those vertices with examples will have small posterior variance, of order $\sigma^2$. This effect disappears again for large $\nu$, where the $O(\sqrt{\nu})$ fluctuations in the number of examples at each vertex becomes relatively small. Mathematically this can be seen from the term proportional to $\nu$ in \eqref{eqn:Opper2}, which for large $\nu$ ensures that only values of $f_i^a$ with $\exp(-\sum_{a}(f_{i}^a)^{2}/2\sigma^{2})$ close to 1 will contribute. A quadratic approximation is then justified even if $\sigma^2$ is not large.

\begin{figure}
\begin{center}
  \input{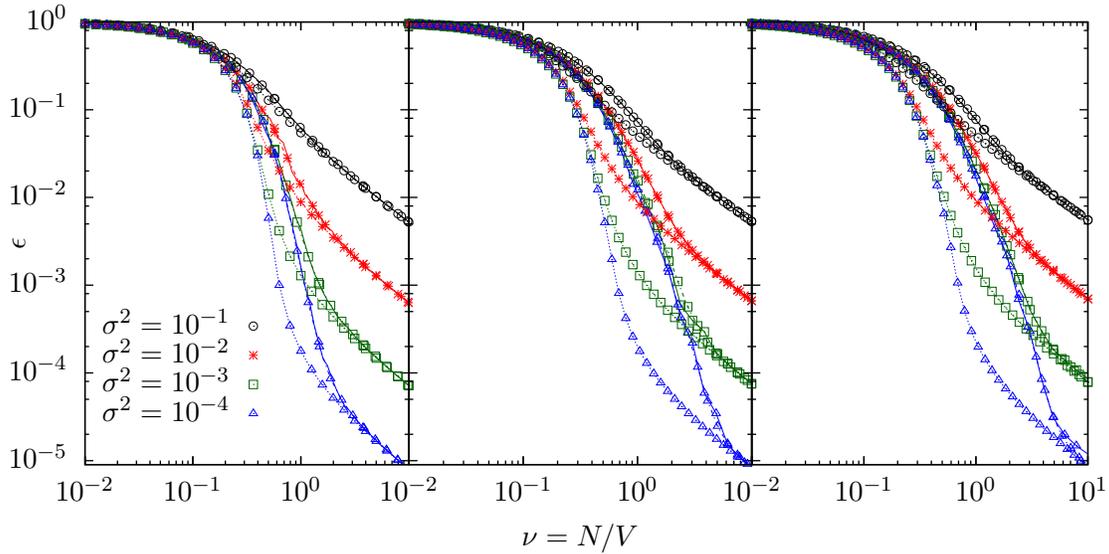}
  \end{center}
  \caption{(Left) Learning curves for GP regression with globally normalised kernels with $p=10$, $a=2$ on 3-regular random graphs for a range of noise levels $\sigma^2$. Dotted lines: eigenvalue predictions (see Section \protect\ref{sec:evalpred}), solid lines: numerically simulated learning
curves for graphs of size $V=500$, dashed lines: cavity predictions (see Section \ref{sec:cavityglobal}); note these are mostly visually indistinguishable from the simulation results. (Centre) As (left)
for Erd\H{o}s-R\'enyi random graphs with mean degree 3. (Right) As (left) for power law generalised random graphs with exponent 2.5 and cutoff 2.\label{fig:erdos} \label{fig:powerlaw}
\label{fig:regular} \label{fig:globallc}}
\end{figure}

Learning curve predictions from Equation~\eqref{eqn:mercerapprox} using numerically computed eigenvalues for the globally normalised random walk kernel are shown in Figure \ref{fig:globallc} as dotted lines for random regular (left), Erd\H{o}s-R\'enyi (centre) and power law generalised random graphs (right). The predictions are compared to numerically simulated learning curves shown as solid lines, for a range of noise levels.
Consistent with the discussion above, the predictions of the eigenvalue approximation are accurate where the Gaussian variational approach is justified, that is, for small and large $\nu$. Figure \ref{fig:globallc} also shows that the accuracy of the approximation improves as the noise level $\sigma^2$ becomes larger, again as expected by the nature of the Gaussian approximation.

\subsubsection{Learning Curves for Large $p$}\label{sec:scaling}

Before moving on to the more accurate cavity prediction of the learning curves, we now look at how the learning curves for GP regression on graphs depend on the kernel lengthscale $p/a$. We focus for this discussion on random regular graphs, where the distinction between global and local normalisation is not important.
In \Sref{sec:dregtree}, we saw that on a large regular graph the random walk kernel approaches a non-trivial limiting form for large $p$, as long as one stays below the threshold \eqref{eqn:p_over_a_limit} for $p$ where cycles become important. One might be tempted to conclude from this that also the learning curves have a limiting form for large $p$. This is too naive however, as one can see by considering, for example, the effect of the first example on the Bayes error. If the example is at vertex $i$, the posterior variance at vertex $j$ is, from \eqref{eqn:GPcovariance}, $C_{jj} - C_{ij}^2/(C_{ii}+\sigma^2)$. As the prior variances $C_{jj}$ are all equal, to unity for our chosen normalisation, this is $1-C_{ij}^2/(1+\sigma^2)$. The reduction in the Bayes error is therefore $\epsilon(0)-\epsilon(1) = (1/V)\sum_{j} C_{ij}^2/(1+\sigma^2)$. As long as cycles are unimportant this is independent of the location of the example vertex $i$, and in the notation of \Sref{sec:dregtree} can be written as
\begin{equation}
\epsilon(0)-\epsilon(1) = \frac{1}{1+\sigma^2} \sum_{l=0}^p v_l C_{l,p}^2,
\label{eqn:initial_error_decay}
\end{equation}
where $v_l$ is, as before, the number of vertices a distance $l$ away from vertex $i$, that is, $v_0=1$, $v_l=d(d-1)^{l-1}$ for $l\geq 1$. To evaluate \eqref{eqn:initial_error_decay} for large $p$, one cannot directly plug in the limiting kernel form \eqref{eqn:clpinfinity}: the `shell volume' $v_l$ just balances the $l$-dependence of the factor $(d-1)^{-l/2}$ from $C_{l,p}$, so that one gets contributions from all distances $l$, proportional to $l^2$ for large $l$. Naively summing up to $l=p$ would give an initial decrease of the Bayes error growing as $p^3$. This is not correct; the reason is that while $C_{l,p}$ approaches the large $p$-limit \eqref{eqn:clpinfinity} for any fixed $l$, it does so more and more slowly as $l$ increases. A more detailed analysis, sketched in Appendix \ref{app:lcscale}, shows that for large $l$ and $p$, $C_{l,p}$ is proportional to the large $p$-limit $l(d-1)^{-l/2}$ up to a characteristic cutoff distance $l$ of order $p^{1/2}$, and decays quickly beyond this. Summing in \eqref{eqn:initial_error_decay} the contributions of order $l^2$ up to this distance predicts finally that the initial error decay should scale, non-trivially, as $p^{3/2}$.

We next show that this large $p$-scaling with $p^{3/2}$ is also predicted, for the entire learning curve, by the eigenvalue approximation \eqref{eqn:mercerapprox}. As before we consider $d$-regular random graphs. The required spectrum of kernel eigenvalues $\lambda_\alpha$ becomes identical, for large $V$, to that on a $d$-regular tree \citep{McKay1981}. Explicitly, if $\lambda_\alpha^L$ are the eigenvalues of the normalised graph Laplacian on a tree, then the kernel eigenvalues are $\lambda_\alpha=\kappa^{-1} V^{-1} (1-\lambda_\alpha^L/a)^p$. Here the factor $V^{-1}$ comes from the same factor in the kernel eigenvalue definition after \eqref{eqn:mercerapprox}, and $\kappa$ is the overall normalisation constant which enforces $\sum_\alpha \lambda_\alpha = V^{-1}\sum_j C_{jj} = 1$. The spectrum of the tree Laplacian is known \citep[see][]{McKay1981,Chung1996} and is given by
\begin{equation}
 \rho(\lambda^L) = \begin{cases}
 \frac{\sqrt{\frac{4(d-1)}{d^{2}}-(\lambda^L-1)^{2}}}{(2\pi /d) \lambda^L(2-\lambda^L)} & \lambda_{-}\leq \lambda \leq \lambda_{+}, \\
 0 & {\rm otherwise},
 \end{cases}
\end{equation}
where $\lambda_{\pm} = 1 \pm \frac{2}{d}(d-1)^{1/2}$. (There are also two isolated eigenvalues at 0 and 2, which do not contribute for large $V$.)

We can now write down the function $g$ from \eqref{eqn:mercerapprox}, converting the sum over kernel eigenvalues to $V$ times an integral over Laplacian eigenvalues for large $V$. Dropping the $L$ superscript, the result is
\begin{equation}\label{eqn:ghregtree}
  g(h) = \int_{\lambda_{-}}^{\lambda_{+}}\rmd\lambda\, \rho(\lambda)[\kappa(1-\lambda/a)^{-p}+hV^{-1}]^{-1}.
\end{equation}
The dependence on $hV^{-1}$ here shows that in the approximate learning curve  \eqref{eqn:mercerapprox}, the Bayes error will depend only on $\nu=N/V$ as might have been expected. The condition for the normalisation factor $\kappa$ becomes simply $g(0)= 1$, or $\kappa^{-1}=\int \rmd\lambda\,\rho(\lambda)(1-\lambda/a)^{p}$.

So far we have written down how one would evaluate the eigenvalue approximation to the learning curve on large $d$-regular random graphs, for arbitrary kernel parameters $p$ and $a$. Now we want to consider the large $p$-limit. We show that there is then a \emph{master curve} for the Bayes error against $\nu p^{3/2}$. This is entirely consistent with the $p^{3/2}$ scaling found above for the initial error decay. The intuition for the large $p$ analysis is that the factor $(1-\lambda/a)^p$ decays quickly as the Laplacian eigenvalue $\lambda$ increases beyond $\lambda_-$, so that only values of $\lambda$ near $\lambda_-$ contribute. One can then approximate
\begin{equation}
  \left(1-\frac{\lambda}{a}\right)^{p}\approx
  \left(1-\frac{\lambda_{-}}{a}\right)^{p}\exp\left(-\frac{p(\lambda-\lambda_{-})}{a-\lambda_{-}}\right).
\end{equation}
Similarly one can replace $\rho(\lambda)$ by its leading square root behaviour near $\lambda_{-}$,
\begin{equation}
  \rho(\lambda) = (\lambda-\lambda_{-})^{1/2}\frac{(d-1)^{1/4}d^{5/2}}{\pi(d-2)^{2}}.
\end{equation}
Substituting these approximations into \eqref{eqn:ghregtree} and introducing the rescaled integration variable $y=p(\lambda-\lambda_{-})/(a-\lambda_{-})$ gives
\begin{equation}
  g(h) = r\kappa^{-1}(1-\lambda_{-}/a)^{p}\left(\frac{a-\lambda_{-}}{p}\right)^{3/2}F(h \kappa^{-1} V^{-1}(1-\lambda_{-}/a)^{p}),
\end{equation}
where $r = (d-1)^{1/4}d^{5/2}/(\pi(d-2)^{2})$ and $F(z)=\int_{0}^{\infty}\rmd y\, y^{1/2}(\exp(y)+z)^{-1}$. Since $g(0)=1$, the prefactor must equal $1/F(0)=2/\sqrt{\pi}$. This fixes the normalisation constant $\kappa$, and we can simplify to
\begin{equation}
  g(h) = \frac{F(hV^{-1}c^{-1})}{F(0)},\qquad c=rF(0)\left(\frac{a-\lambda_{-}}{p}\right)^{3/2}.
\end{equation}
The learning curves for large $p$ are then predicted from \eqref{eqn:mercerapprox} by solving
\begin{equation}
\epsilon = F(\nu c^{-1}/(\epsilon+\sigma^{2}))/F(0),
\label{eqn:mastercurve}
\end{equation}
and depend clearly only on the combination $\nu c^{-1}$. Because $c$ is proportional to $p^{-3/2}$, this shows that learning curves for different $p$ should collapse onto a master curve when plotted against $\nu p^{3/2}$.

\begin{figure}
  \input{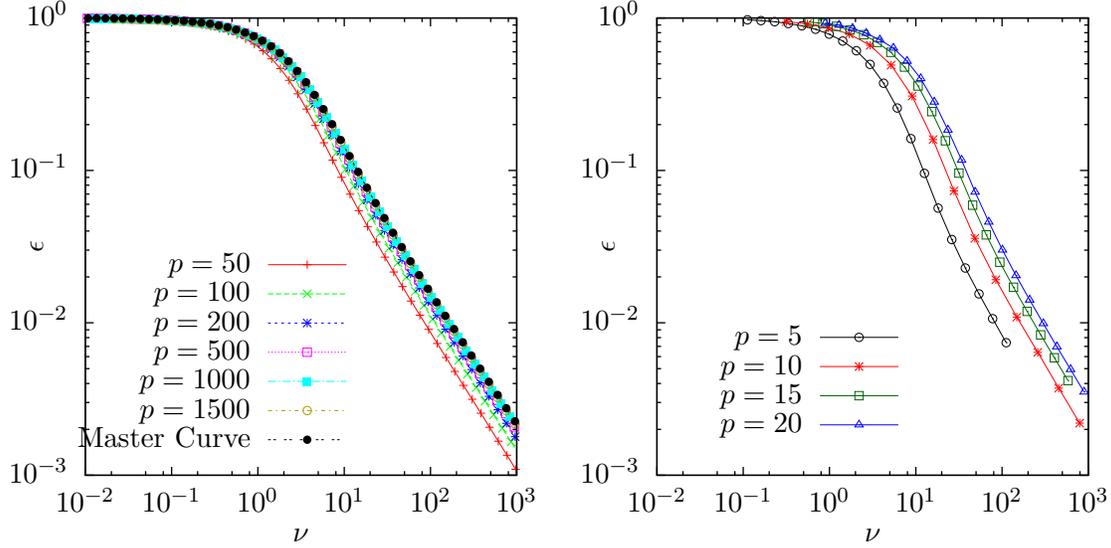}
  \caption{(Left) Eigenvalue approximation for learning curves on a random 3-regular graph, using a random walk kernel with $a=2$, $\sigma^{2}=0.1$ and increasing values of $p$ as shown. Plotting against $\nu p^{3/2}$ shows that for large $p$ these rescaled curves approach the master curve predicted from \protect\eqref{eqn:mastercurve}, though this approach is slower in the tail of the curves. (Right) As (left), but for numerically simulated learning curves on graphs of size $V=500$.\label{fig:taillcscale}\label{fig:cavlcscale}}
\end{figure}

A plot of the scaling of the eigenvalue learning curve approximations onto the master curve is shown in Figure \ref{fig:taillcscale} (left). As can be seen, large values of $p$ are required in order to get a good collapse in the tail of the learning curve prediction, whereas in the initial part the $p^{3/2}$ scaling is accurate already for relatively small $p$.

Finally, Figure \ref{fig:taillcscale} (right) shows that the predicted $p^{3/2}$-scaling of the learning curves is present not only within the eigenvalue approximation, but also in the actual learning curves. Figure \ref{fig:cavlcscale} (right) displays numerically simulated learning curves for $p=5,10,15$ and $20$, against the rescaled number of examples $\nu p^{3/2}$ as before. Even for these comparatively small values of $p$ one sees that the rescaled learning curves approach a master curve.

\section{Exact Learning Curves: Cavity Method}\label{sec:cavitypred}

So far we have discussed the eigenvalue approximation of GP learning curves, and how it deviates from numerically exact simulated learning curves.
As discussed in \Sref{sec:evalpred}, the deficiencies of the eigenvalue approximation can be traced back to the fact that the fluctuations in the number of training examples seen at each vertex of the graph cannot be accounted for in detail. If in the average over data sets these fluctuations could be treated exactly, one would hope to obtain exact, or at least very accurate, learning curve predictions.
In this section we show that this is indeed possible in the case of a random walk kernel, for both global and local normalisations. We derive our prediction using belief propagation or, equivalently, the cavity method \citep{Mezard2003}. The approach relies on the fact that the local structure of the graph on which we are learning is tree-like. This local tree-like structure always occurs in large random graphs sampled uniformly from an ensemble specified by an arbitrary but fixed degree distribution, which is the scenario we consider here. We will see that already for moderate graph sizes of $V=500$, our predictions are nearly indistinguishable from numerical simulations.

In order to apply the cavity method to the problem of predicting learning curves we must first rewrite the partition function \eqref{eqn:Z} in the form of a graphical model. This means that the function being integrated over to obtain $Z$ must consist of factors relating only to individual vertices, or to pairs of neighbouring vertices. The inverse of the covariance matrix in \eqref{eqn:Z} creates factors linking vertices at arbitrary distances along the graph, and so must be eliminated before the cavity method can be applied. We begin by assuming a general form for the normalisation of $\hat{C}$ that encompasses both local and global normalisation and set $\bm{C}=\bm{\mathcal{K}}^{-1/2} [(1-a^{-1})\bm{I} + a^{-1}\bm{D}^{-1/2}\bm{A}\bm{D}^{-1/2}]^{p}\bm{\mathcal{K}}^{-1/2}$ with $\mathcal{K}_{ij} = \kappa_i\delta_{ij}$. To eliminate interactions across the entire graph we first Fourier transform the prior term $\exp(-\frac{1}{2}\f\T\C^{-1}\f)$ in \eqref{eqn:Z}, introduce Fourier variables $\bm{h}$, and then integrate out the remaining terms with respect to $\f$ to give
\begin{equation}\label{eqn:Zfourier}
  Z \propto \prod_{i}\left(\frac{\ni[i]}{\sigma^{2}}+\lambda\right)^{-1/2}\int \rmd\bm{h}\exp\left(-\frac{1}{2}\bm{h}\T\bm{C}\bm{h} -\frac{1}{2}\bm{h}\T\textrm{diag}\left(\left(\frac{\ni[i]}{\sigma^{2}}+\lambda\right)^{-1/2}\right)\bm{h}\right).
\end{equation}
The coupling between different vertices in (\ref{eqn:Zfourier}) is now through $\bm{C}$ so still links vertices up to distance $p$. To reduce these remaining interactions to ones among nearest neighbours only, one exploits the binomial expansion of the random walk kernel given in \eqref{eqn:binomrandomwalk}. Defining $p$ additional variables at each vertex as $\bm{h}^{q} =
\bm{\mathcal{K}}^{1/2}(\bm{D}^{-1/2}\bm{A}\bm{D}^{-1/2})^q\bm{\mathcal{K}}^{-1/2}\bm{h}$, $q=1,\ldots,p$, and abbreviating
$c_q = \binom{p}{q}(1-a^{-1})^{p-q}(a^{-1})^q$, the interaction term $\bm{h}\T\bm{C}\bm{h}$ turns into a local term $\sum_{q=0}^{p}c_{q}(\bm{h}^0)\T\bm{\mathcal{K}}^{-1}\bm{h}^q$.
(Here we have, for the sake of uniformity, written $\bm{h}^0$ instead of $\bm{h}$.)
Of course the interactions have only been `hidden' in the $\bm{h}^q$, but the key point is that the definition of these additional variables can be enforced recursively, via $\bm{h}^{q} = \bm{\mathcal{K}}^{1/2}\bm{D}^{-1/2}\bm{A}\bm{D}^{-1/2}\bm{\mathcal{K}}^{-1/2}\bm{h}^{q-1}$. We represent this definition via a Dirac delta function (for each $q=1,\ldots,p$) and then Fourier transform the latter, with conjugate variables $\bm{\hat{h}}^{q}$, to get
\begin{multline}\label{eqn:Zglobalbinom}
    Z \propto \prod_{i}\left(\frac{\ni[i]}{\sigma^{2}}+\lambda\right)^{-1/2}\int \prod_{q=0}^{p}\rmd\bm{h}^{q}\prod_{q=1}^{p}\rmd\bm{\hat{h}}^{q}\exp\Bigg(-\frac{1}{2}(\bm{h}^0)\T\textrm{diag}\left(\left(\frac{\ni[i]}{\sigma^{2}}+\lambda\right)^{-1}\right)\bm{h}^0\\
    -\frac{1}{2}\sum_{q=0}^{p}c_{q}(\bm{h}^0)\T\bm{\mathcal{K}}^{-1}\bm{h}^q +\rmi\sum_{q=1}^p(\bm{\hat{h}}^{q})\T\left(\bm{h}^{q} - \bm{\mathcal{K}}^{1/2}\bm{D}^{-1/2}\bm{A}\bm{D}^{-1/2}\bm{\mathcal{K}}^{-1/2}\bm{h}^{q-1}\right)\Bigg).
\end{multline}
Because the graph adjacency matrix $\bm{A}$ now appears at most linearly in the exponent, all interactions are between nearest neighbours only. We have thus expressed our $Z$ as the partition function of a (complex-valued) graphical model.

\subsection{Global Normalisation}\label{sec:cavityglobal}
We can now apply belief propagation to the calculation of marginals for the above graphical model. We focus first on the simpler case of a globally normalised kernel where $\kappa_i = \kappa$ for all $i$. Rescaling each $h_{i}^{q}$ to $d_{i}^{1/2}\kappa^{1/2}h_{i}^{q}$ and $\hat{h}_{i}^{q}$ to $d_{i}^{1/2}\hat{h}_{i}^{q}/\kappa^{1/2}$ we are left with
\begin{multline}\label{eqn:Zglobalsite}
    Z\propto \prod_{i}\left(\frac{\ni[i]}{\sigma^{2}}+\lambda\right)^{-1/2}\int \prod_{q=0}^{p}\rmd\bm{h}^{q}\prod_{q=1}^{p}\rmd\bm{\hat{h}}^{q}\prod_{i}\exp\left(-\frac{1}{2}\sum_{q=0}^{p}c_{q}h_{i}^{0}h_{i}^{q}d_{i} - \frac{1}{2}\frac{(h_{i}^{0})^{2}\kappa d_{i}}{\ni[i]/\sigma^{2}+\lambda} +\rmi\sum_{q=1}^{p}d_{i}\hat{h}^q_{i}h_{i}^{q}\right)\\
    \prod_{(i,j)}\exp\left(-\rmi\sum_{q=1}^{p}\left(\hat{h}_{i}^{q}h_{j}^{q-1} + \hat{h}_{j}^{q}h_{i}^{q-1}\right)\right),
\end{multline}
where the interaction terms coming from the adjacency matrix, $\bm{A}$, have been written explicitly as a product over distinct graph edges $(i,j)$.

To see how the Bayes error (\ref{eqn:bayeserror}) can be obtained from this partition function, we differentiate $\log(Z)$ with respect to $\lambda$ as prescribed by \eqref{eqn:epgZ} to get
\begin{equation}\label{eqn:globalepg}
  \epsilon(\nu) =
\lim_{\lambda\to0}\frac{1}{V}\sum_{i}\frac{1}{\ni[i]/\sigma^{2}+\lambda}
\left(1-\frac{d_i\kappa\langle(h_{i}^{0})^{2}\rangle}{\ni[i]/\sigma^{2}+\lambda}\right).
\end{equation}
In order to calculate the Bayes error we therefore require specifically the marginal distributions of $h_i^0$. These can be calculated using the cavity method: for a large random graph with arbitrary fixed degree sequence the graph is locally tree-like, so that if
vertex $i$ were eliminated the corresponding subgraphs (locally trees) rooted at the neighbours $j\in\mathcal{N}(i)$ of
$i$ would become approximately independent. The resulting cavity marginals created by removing $i$, which we denote $P^{(i)}_{j}(\bm{h}_j,\bm{\hat{h}}_{j}|\x)$, can then be calculated
iteratively within these subgraphs using the update equations
\begin{multline}\label{eqn:globalcavity}
  P^{(i)}_{j}(\bm{h}_j,\bm{\hat{h}}_{j}|\x) \propto \exp\left(-\frac{1}{2}\sum_{q=0}^{p}c_{q}d_{j}h^{0}_{j}h_{j}^{q}
-\frac{1}{2}\frac{d_{j}\kappa(h_{j}^{0})^{2}}{\ni[j]/\sigma^{2}+\lambda}
+ \rmi\sum_{q=1}^{p}
d_j\hat{h}^{q}_{j}h_{j}^q\right)\\
\int\prod_{k\in\mathcal{N}(j)\backslash i} \rmd\bm{h}_{k}\rmd\bm{\hat{h}}_{k}\,
\exp\left(-\rmi\sum_{q=1}^{p}(\hat{h}_{j}^{q}h_{k}^{q-1} +
\hat{h}_{k}^{q}h_{j}^{q-1})\right)P_{k}^{(j)}(\bm{h}_k,\bm{\hat{h}}_k|\x).
\end{multline}
where $\bm{h}_j = (h_{j}^0,\ldots,h_{j}^p)\T$ and $\bm{\hat{h}}_j = (\hat{h}_{j}^1,\ldots,\hat{h}_{j}^p)\T$. In terms of the sum-product formulation of belief propagation, the cavity marginal on the left is the message that vertex $j$ sends to the factor in $Z$ for edge $(i,j)$ \citep{Bishop2007}.

One sees that the cavity update Equations~(\ref{eqn:globalcavity}) are solved self-consistently by complex-valued Gaussian distributions with mean zero and
covariance matrices $\bm{V}_{j}^{(i)}$. This Gaussian character of the solution was of course to be expected because in \eqref{eqn:Zglobalsite} we have a Gaussian graphical model. By performing the Gaussian integrals in the cavity update equations explicitly, one finds for the corresponding updates of the covariance matrices the rather simple form
\begin{equation}
\bm{V}_{j}^{(i)}= (\bm{O}_{j}
-\sum_{k\in\mathcal{N}(j)\backslash i}\bm{X}\bm{V}_{k}^{(j)}\bm{X})^{-1},
\label{eqn:globalvariance}
\end{equation}
where we have defined the $(2p+1)\times(2p+1)$ matrices
\begin{equation}
\setlength{\arraycolsep}{1mm}
  \bm{O}_j = d_{j}\left(\begin{array}{cccc|ccc}
    c_0 \!+\!\frac{\kappa}{\ni[j]/\sigma^{2} +\lambda} &
\frac{c_1}{2} & \dots & \frac{c_p}{2} &
0 & \dots & 0 \\
\frac{c_1}{2}& & & &
-\rmi & & \\
\vdots & & & &
 & \ddots & \\
\frac{c_p}{2}& & & &
 & & -\rmi\\[0.5mm]
\hline
0 & -\rmi & & &
 & & \\
\vdots & & \ddots & &
 & \bm{0}_{p,p} & \\
0 & & & -\rmi &
 & &
\end{array}\right), \quad
\bm{X} = \left(\begin{array}{cccc|ccc}
 & & & & \rmi & & \\
 & \multicolumn{2}{c}{\bm{0}_{p+1,p+1}} & & & \ddots & \\
 & & & & & & \rmi\\
 & & & & 0 & \dots & 0\\
\hline
\rmi & & & 0 & & & \\
  & \ddots & & \vdots & & \bm{0}_{p,p}\\
 & & \rmi & 0 & & &
\end{array}\right).\label{eqn:globalOX}
\end{equation}

At first glance \eqref{eqn:globalvariance} becomes singular for $\ni[j] = 0$; however this is easily avoided. We introduce $\bm{O}_j-\sum_{k=1}^{d-1}\bm{X}\bm{V}^{(j)}_k\bm{X}=\bm{M}_{j}+ [d_j\kappa/(\ni[j]/\sigma^{2}+\lambda)]\bm{e}_{0}\bm{e}_{0}^{T}$ with $\bm{e}_0^{T}=(1,0,\ldots,0)$ so that $\bm{M}_j$ contains all the non-singular terms. We may then apply the Woodbury identity \citep{Hager1989} to write the matrix inverse in a form where the $\lambda\to 0$ limit can be taken without difficulties:
\begin{equation}
\left(\bm{O}_j-\sum_{k=1}^{d-1}\bm{X}\bm{V}^{(j)}_k\bm{X}\right)^{-1} = \bm{M}_{j}^{-1} - \frac{\bm{M}_{j}^{-1}\bm{e}_{0}\bm{e}_{0}^{T}\bm{M}_{j}^{-1}}{(\ni[j]/\sigma^{2}+\lambda)/(d_j\kappa) +\bm{e}_{0}^{T}\bm{M}_{j}^{-1}\bm{e}_{0}}.
\end{equation}

In our derivation so far we have assumed a fixed graph, we therefore need to translate these equations to the setting we ultimately want to study, that is, an ensemble of large random graphs. This ensemble is characterised by
the distribution $p(d)$ of the degrees $d_i$, so that every graph that has the desired degree distribution is assigned
equal probability. Instead of individual cavity covariance matrices $\bm{V}_{j}^{(i)}$, one must then consider their
probability distribution $W(\bm{V})$ across all edges of the graph. Picking at random an edge $(i,j)$ of a graph, the
probability that vertex $j$ will have degree $d_j$ is then $p(d_j)d_j/\bar{d}$, because such a vertex has $d_j$
`chances' of being picked. (The normalisation factor is the average degree $\bar{d}=\sum_i p(d_i)d_i$.) Using again the locally
treelike structure, the incoming (to vertex $j$) cavity covariances $\bm{V}_k^{(j)}$ will be independent and identically distributed samples from
$W(\bm{V})$. Thus a fixed point of the cavity update equations corresponds to a fixed point of an update equation for
$W(\bm{V})$:
\begin{equation}\label{eqn:globalensembleupdate}
W(\bm{V}) = \sum_{d}\frac{p(d)d}{\bar{d}}\left\langle\int
\prod_{k=1}^{d-1} \rmd\bm{V}_k\, W(\bm{V}_k)\
\delta\left(\bm{V} -\left(\bm{O} -  \sum_{k=1}^{d-1}\bm{X}\bm{V}_{k}\bm{X}\right)^{-1}\right)\right\rangle_{\ni}.
\end{equation}
Since the vertex label is now arbitrary, we have omitted the index $j$. The average in \eqref{eqn:globalensembleupdate} is over the distribution of the number of examples $\ni\equiv
\ni[j]$ at vertex $j$. As before we assume for simplicity that examples are drawn with uniform input probability
across all vertices, so that the distribution of $\ni$ is simply \mbox{$\ni\sim\textrm{Poisson}(\nu)$} in the limit of large $N$ and $V$ at fixed $\nu=N/V$.

In general Equation~\eqref{eqn:globalensembleupdate}---which can also be formally derived using the replica approach
\citep[see][]{Urry2012}---cannot be solved analytically, but we can tackle it numerically using
population dynamics \citep{Mezard2001}. This is an iterative technique where one creates a population of
covariance matrices and for each iteration updates a random element of
the population according to the delta function in
\eqref{eqn:globalensembleupdate}. The update is calculated
by sampling from the degree distribution $p(d)$ of local degrees, the Poisson distribution of the local number of
examples $\nu$ and from the distribution $W(\bm{V}_k)$ of `incoming'
covariance matrices, the latter being approximated
by uniform sampling from the current population.

Once we have $W(\bm{V})$, the Bayes error can be found from the graph ensemble version of Equation~\eqref{eqn:epgZ}. This is obtained by inserting the explicit expression for
$\langle (h_i^0)^2\rangle$ in terms of the cavity marginals of the neighbouring vertices, and replacing the average over vertices with an average over degrees $d$:
\begin{equation}
\label{eqn:epgglobalcavity}
\epsilon(\nu) = \lim_{\lambda\to 0}
\sum_{d}p(d)\left\langle
\frac{1}{\ni/\sigma^{2}+\lambda}
\left(1-\frac{d\kappa}{\ni/\sigma^{2}+\lambda}
\int \prod_{k=1}^{d} \rmd\bm{V}_k\, W(\bm{V}_k)\
(\bm{O} - \sum_{k=1}^{d}\bm{X}\bm{V}_{k}\bm{X})^{-1}_{00}
\right)\right\rangle_{\ni}.
\end{equation}
The number of examples at the vertex $\ni$ is once more to be averaged over $\ni\sim\textrm{Poisson}(\nu)$. The subscript `00'
indicates the top left element of the matrix, which determines the variance of $h^0$.

To be able to use Equation~\eqref{eqn:epgglobalcavity}, we again need to rewrite it into a form that remains explicitly
non-singular when $\ni=0$ and $\lambda\to 0$. We separate the $\ni$-dependence of the matrix inverse again and write, in slightly modified notation as appropriate for the graph ensemble case,
$\bm{O}-\sum_{k=1}^d\bm{X}\bm{V}_k\bm{X}=\bm{M}_d+ [d\kappa/(\ni/\sigma^{2}+\lambda)]\bm{e}_{0}\bm{e}_{0}^{T}$, where
$\bm{e}_0^{T}=(1,0,\ldots,0)$. The $00$ element of the matrix inverse appearing above can then be expressed using the Woodbury formula \citep{Hager1989} as
\begin{equation}\label{eqn:wood}
\e0\T\left(\bm{O}-\sum_{k=1}^d\bm{X}\bm{V}_k\bm{X}\right)^{-1}\!\!\!\!\!\!\e0 = \e0\T\bm{M}_d^{-1}\e0 - \frac{\e0\T\bm{M}_d^{-1}\bm{e}_{0}\bm{e}_{0}^{T}\bm{M}_d^{-1}\e0}{(\ni/\sigma^{2}+\lambda)/(d\kappa) +\bm{e}_{0}^{T}\bm{M}_d^{-1}\bm{e}_{0}}.
\end{equation}
The $\lambda\to0$ limit can now be taken, with the result
\begin{equation}\label{eqn:epgglobalwood}
\epsilon(\nu) = \left\langle
\sum_{d}p(d)
\int \prod_{k=1}^{d} \rmd\bm{V}_k\, W(\bm{V}_k)\
\frac{1}{\ni/\sigma^2 + d\kappa(\bm{M}_d^{-1})_{00}}\right\rangle_{\ni}.
\end{equation}
This has a simple interpretation: the cavity marginals of the neighbours provide an effective Gaussian prior for each
vertex, whose inverse variance is $d\kappa(\bm{M}^{-1})_{00}$.

The self-consistency Equation~\eqref{eqn:globalensembleupdate} for $W(\bm{V})$ and the expression
\eqref{eqn:epgglobalwood} for the resulting Bayes error allow us to predict learning curves as a function of the number
of examples per vertex, $\nu$, for \emph{arbitrary degree distributions} $p(d)$ of our random graph ensemble. For large graphs the predictions should become exact. It is worth stressing that such exact learning curve predictions have previously only been available in very specific, noise-free, GP regression scenarios, while our result for GP regression on graphs is applicable to a broad range of random graph ensembles, with arbitrary noise levels and kernel parameters.

We note briefly that for graphs with isolated vertices ($d=0$), one has to be slightly careful: already in the
definition of the covariance function \eqref{eqn:randomwalkkernel} one should replace $\bm{D}\to \bm{D}+\delta \bm{I}$
to avoid division by zero, taking $\delta\to0$ at the end. For $d=0$ one then finds in the expression
\eqref{eqn:epgglobalwood} that $(\bm{M}^{-1})_{00}=1/(c_{0}\delta)$, where $c_0$ is defined before \eqref{eqn:Zglobalbinom}.
As a consequence, $\kappa(\delta+d)
(\bm{M}^{-1})_{00}=\kappa\delta (\bm{M}^{-1})_{00}=\kappa/c_0$. This is to be expected since isolated vertices each have a separate Gaussian prior with variance $c_0/\kappa$.

Equations \eqref{eqn:globalensembleupdate} and \eqref{eqn:epgglobalwood} still require the normalisation constant, $\kappa$. The simplest way to calculate this is to run the population dynamics once for $\kappa=1$ and $\nu=0$, that is, an unnormalised kernel and no training data. The result for $\epsilon$ then just gives the average (over vertices) prior variance. With $\kappa$ set to this value, one can then run the population dynamics for any $\nu$ to obtain the Bayes error prediction for GP regression with a globally normalised kernel.

Comparisons between the cavity prediction for the learning curves, numerically exact simulated learning curves and the results of the eigenvalue approximation are shown in Figure
\ref{fig:globallc} (left, centre and right), for regular, Erd\H{o}s-R\'enyi and generalised random graphs with power law degree distributions respectively. As can be
seen the cavity predictions greatly outperform the eigenvalue approximation and are accurate along the whole length of the curve. This confirms our expectation that the cavity approach will become exact on large graphs, although it is remarkable that the agreement is quantitatively so good already for graphs with only five hundred vertices.

\subsection{Predicting Prior Variances}\label{sec:cavityvariance}
As a by-product of the cavity analysis for globally normalised kernels we note that in the cavity form
of the Bayes error in Equation~\eqref{eqn:epgglobalwood}, the fraction
$(\ni/\sigma^2 + d\kappa(\bm{M}_d^{-1})_{00})^{-1}$ is the local Bayes error, that is, the local posterior variance. By keeping track of individual samples for this quantity from the population dynamics approach, we can thus predict the distribution of local posterior variances. If we set $\nu=0$, then this becomes the distribution of prior variances. The cavity approach therefore gives us, without additional effort, a prediction for this distribution.

We can now go back to \Sref{sec:kernelnorm} and compare the cavity predictions to numerically simulated distributions of prior variances. The cavity predictions for these distributions are shown by the black lines in Figure \ref{fig:priorvariance}. The cavity approach provides, in particular, detailed information about the tail of the distributions as shown in the insets. There is good agreement between the predictions and the numerical simulations, both regarding the general shape of the variance distributions and the fine structure with a number of non-trivial peaks and troughs. The residual small shifts between the predictions and the numerical results
for a single instance of a large graph are most likely due to finite size effects: in a finite graph, the assumption
of a tree-like local structure is not exact because there can be rare short cycles; also, the long cycles that the cavity method ignores because their length diverges logarithmically with $V$ will have an effect when $V$ is finite.

\subsection{Local Normalisation}\label{sec:cavitylocal}

We now extend the cavity analysis for the learning curves to the case of locally normalised random walk kernels, which, as argued above, provide more plausible probabilistic models. In this case the diagonal entries of the normalisation matrix $\bm{\mathcal{K}}$ are defined as
\begin{equation}
  \kappa_i = \int \rmd\bm{f}f_{i}^{2}P(\bm{f}),
\end{equation}
where $P(\f)$ is the GP prior with the unnormalised kernel $\bm{\hat{C}}$. This makes clear why the locally normalised kernel case is more challenging technically: we cannot calculate the normalisation constants once and for all for a given random graph ensemble and set of kernel parameters $p$ and $a$ as we did for $\kappa$ in the globally normalised scenario. Instead we have to account for the dependence of the $\kappa_i$ on the specific graph instance.

On a single graph instance, this stumbling block can be overcome as follows. One iterates the cavity updates \eqref{eqn:globalvariance} for the unnormalised kernel and without the training data (i.e., setting $\kappa=1$ and $\ni[i]=0$). The local Bayes error at vertex $i$, given by the $i$-th term in the sum from \eqref{eqn:globalepg}, then gives us $\kappa_i$. Because $\ni[i]=0$, one has to use the Woodbury trick to get well-behaved expressions in the limit where the auxiliary parameter $\lambda\to 0$, as explained after \eqref{eqn:epgglobalcavity}.

Once the $\kappa_i$ have been determined in this way, one can use them for predicting the Bayes error for the scenario we really want to study, that is, using a locally normalised kernel and incorporating the training data.
%
%If one instead chooses to normalise using more plausible local normalisation then local normalisation factors are tied to local graph structure. One therefore cannot, as we have done for global normalisation, simply calculate the marginals for a single graph case and generalise to ensembles. This apparent stumbling block however can be solved by introducing a second set of cavity equations to calculate the local normalisation factors.
%
The relevant partition function is the analogue of \eqref{eqn:Zglobalbinom} for local normalisation. Dropping the prefactors, the resulting $Z$ can be written as
\begin{multline}\label{eqn:Zlocalfinal}
Z \propto \int \prod_{q=0}^p
\rmd\bm{h}^{q}\prod_{q=1}^p \rmd\bm{\hat{h}}^{q}\prod_{i}\exp\left(-\frac{1}{2}\sum_{q=0}^{p}c_{q}d_{i}h^{0}_{i}h_{i}^{q}
-\frac{1}{2}\frac{d_{i}\kappa_{i}(h_{i}^{0})^{2}}{\ni[i]/\sigma^{2}+\lambda}
+ \rmi\sum_{q=1}^pd_i\hat{h}^{q}_{i}h_{i}^q\right)\\
\prod_{(i,j)}\exp\left(-\rmi\sum^p_{q=1}(\hat{h}_{i}^{q}h_{j}^{q-1} +
\hat{h}_{j}^{q}h_{i}^{q-1})\right),
\end{multline}
where we have rescaled $h_{i}^{q}$ to $d_{i}^{1/2}\kappa_{i}^{1/2}h_{i}^{q}$ and $\hat{h}_{i}^{q}$ to
$d_{i}^{1/2}\kappa_{i}^{-1/2}\hat{h}_{i}^{q}$. Given that the $\kappa_i$ have already been determined, this is a graphical model for which marginals can be calculated by iterating to a fixed point the equations for the cavity marginals:
\begin{multline}\label{eqn:localcavity}
  P_{\textrm{loc},j}^{(i)}(\bm{h}_j,\bm{\hat{h}}_{j}|\x) \propto \exp\left(-\frac{1}{2}\sum_{q=0}^{p}c_{q}d_{j}h^{0}_{j}h_{j}^{q}
  -\frac{1}{2}\frac{d_{j}\kappa_{j}(h_{j}^{0})^{2}}{\ni[j]/\sigma^{2}+\lambda}
+ \rmi\sum_{q=1}^{p}
d_j\hat{h}^{q}_{j}h_{j}^q\right)\\
\int\prod_{k\in\mathcal{N}(j)\backslash i} \rmd\bm{h}_{k}\rmd\bm{\hat{h}}_{k}\,
\exp\left(-\rmi\sum_{q=1}^{p}(\hat{h}_{j}^{q}h_{k}^{q-1} +
\hat{h}_{k}^{q}h_{j}^{q-1})\right)P_{\textrm{loc},k}^{(j)}(\bm{h}_k,\bm{\hat{h}}_k|\x).
\end{multline}
As in Section \ref{sec:cavityglobal} these update equations are solved by cavity marginals of complex Gaussian form, and so we can simplify them to updates for the covariance matrices:
\begin{equation}
  \bm{V}_{\textrm{loc},j}^{(i)}= \left(\bm{O}_{\textrm{loc},j}
  -\sum_{k\in\mathcal{N}(j)\backslash i}\bm{X}\bm{V}_{k,\textrm{loc}}^{(j)}\bm{X}\right)^{-1}.
\label{eqn:localvariance}
\end{equation}
Here $\bm{X}$ is defined as in Equation~\eqref{eqn:globalOX} and $\bm{O}_{\textrm{loc},j}$ is the obvious analogue of
$\bm{O}_j$ also defined in Equation~\eqref{eqn:globalOX}; specifically, $\kappa$ is replaced by $\kappa_j$. Once the update equations have converged, one can calculate the Bayes error from a similarly adapted version of \eqref{eqn:globalepg}.

The above procedure for a single fixed graph now has to be extended to the case of an ensemble of large random graphs characterised by some degree distribution $p(d)$. The outcome of the first round of cavity updates, for the unnormalised kernel without training data, is then represented by a distribution of cavity covariances $\bm{V}$, while the second one gives a distribution of cavity covariances $\bm{V}_{\textrm{loc}}$ for the locally normalised kernel, with training data included. Importantly, these message distributions are coupled to each other via the graph structure, so we need to look at the joint distribution
$W(\bm{V}_{\textrm{loc}},\bm{V})$.

Detailed analysis using the replica method \citep{Urry2012} shows that the correct fixed point equation updates the $\bm{V}$-messages as in the globally normalised case with $\ni=0$. The second set of local covariances, $\bm{V}_{{\rm loc}}$, are then updated according to \eqref{eqn:localvariance}, with a normaliser calculated using the marginals from the $d-1$ $\bm{V}$-covariances and an additional `counterflow' covariance generated from $W(\bm{V}) = \int d\bm{V}_{\textrm{loc}}W(\bm{V}_{\textrm{loc}},\bm{V})$, subject to the constraint that the local marginals of the neighbours are consistent. We find in practice that the consistency constraint can be dropped and the fixed point equation for the distribution of the two sets of messages can be approximated by
\begin{multline}\label{eqn:localensembleupdate}
  W(\bm{V}_{\textrm{loc}},\bm{V}) =\left\langle\sum_{d}\frac{p(d)d}{\bar{d}}\int
  \prod_{k=1}^{d-1} \rmd\bm{V}_k\rmd\bm{V}_{\textrm{loc},k}\rmd\bm{V}_{d}\, \prod_{k=1}^{d-1}W(\bm{V}_{\textrm{loc},k},\bm{V}_{k})W(\bm{V}_{d})\right.\\
  \left.\delta\left(\bm{V}_{\textrm{loc}} -\left(\bm{O}_{\textrm{loc},j}
  -\sum_{k=1}^{d-1}\bm{X}\bm{V}_{k,\textrm{loc}}^{(j)}\bm{X}\right)^{-1}\right)
\delta\left(\bm{V} -\left(\bm{O} -  \sum_{k=1}^{d-1}\bm{X}\bm{V}_{k}\bm{X}\right)^{-1}\right)\right\rangle_{\ni}.
\end{multline}
One sees that if one marginalises over $\bm{V}_{\textrm{loc}}$, then one obtains exactly the same condition on $W(\bm{V})$ as before in the globally normalised kernel case (but with $\kappa=1$ and $\nu=0$), see \eqref{eqn:globalensembleupdate}. This reflects the fact that the cavity updates for the first set of messages on a single graph do not rely on any information about the second set.
The first delta function in \eqref{eqn:localensembleupdate} corresponds to the fixed point condition for this second set of cavity updates. This condition depends, via the value of the local $\kappa$, on the  $\bm{V}$-cavity covariances:
\begin{equation}\label{eqn:kappa_i}
\kappa = \frac{1}{d(\bm{M}_d^{-1})_{00}}.
\end{equation}
It may seem unusual that $d$ copies of $\bm{V}$ enter here; $\bm{V}_d$ represents the cavity covariance from the first set that is {\em received} from the vertex to which the new message $\bm{V}_{\textrm{loc}}$ is being {\em sent}. While this counterflow appears to run against the basic construction of the cavity or belief propagation method, it makes sense here because the first set of cavity messages (or equivalently the distribution $W(\bm{V})$) reaches a fixed point that is independent of the second set, so the counterflow of information is only apparent. The reason why knowledge about $\bm{V}_d$ is needed in the update is that $\kappa$ is the variance of a full marginal rather than a cavity marginal.

Similarly to the case of global normalisation, \eqref{eqn:localensembleupdate} can be solved by looking for a fixed point of $W(\bm{V}_{\textrm{loc}},\bm{V})$ using population dynamics. Updates are made by first updating $\bm{V}$ using Equation~\eqref{eqn:globalcavity} and then updating $\bm{V}_{\textrm{loc}}$ using \eqref{eqn:localcavity} with $\kappa \equiv\kappa_i$ replaced by \eqref{eqn:kappa_i}.

Once a fixed point has been calculated for the covariance distribution we apply the Woodbury formula to \eqref{eqn:globalepg} in a similar manner to Section \ref{sec:cavityglobal} to give the prediction for the learning curve for GP regression with a locally normalised kernel. The result for the Bayes error becomes
\begin{equation}\label{eqn:epglocalwood}
\epsilon = \left\langle
\sum_{d}p(d)
\int \prod_{k=1}^{d} \rmd\bm{V}_{\textrm{loc},k}\rmd\bm{V}_k\, \prod_{k=1}^{d}W(\bm{V}_{\textrm{loc},k},\bm{V}_k)\
\frac{1}{\ni/\sigma^2 + (\bm{M}_{d,\textrm{loc}}^{-1})_{00}/(\bm{M}_d^{-1})_{00}}\right\rangle_{\ni}.
\end{equation}

\begin{figure}
  \input{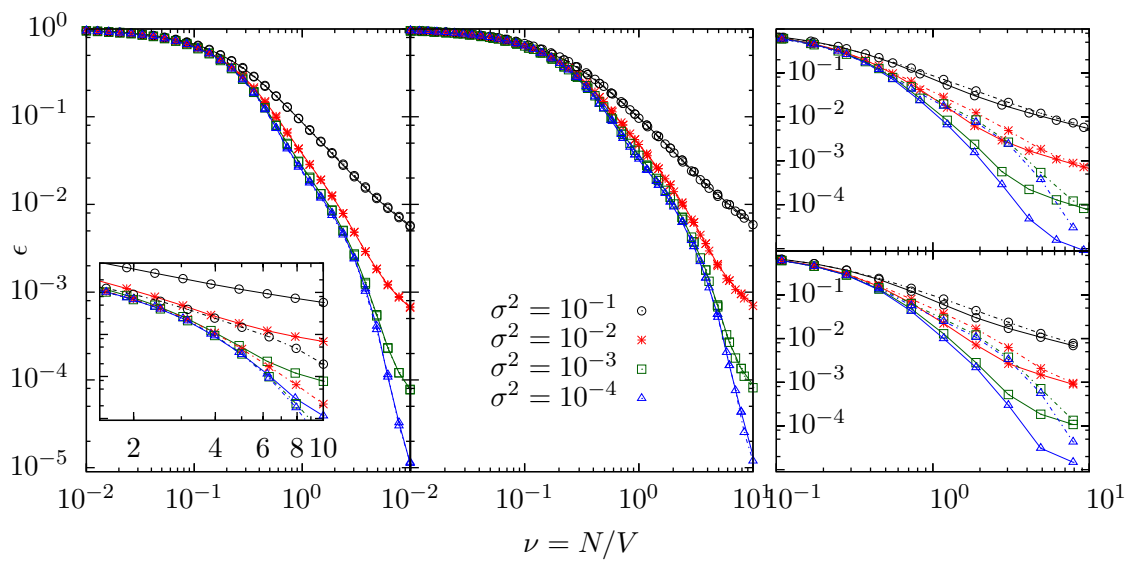}
  \caption{(Left) Learning curves for GP regression with locally normalised kernels with $p=10$, $a=2$ on Erd\H{o}s-R\'enyi random graphs with mean degree 3, for a range of noise levels $\sigma^2$. Solid lines: numerically simulated learning
curves for graphs of size $V=500$, dashed lines: cavity predictions (see Section \protect\ref{sec:cavityglobal}); note these are mostly visually indistinguishable from the simulation results.
(Left inset) Dotted lines
show single vertex contributions to the learning curve (solid line). (Centre) As (left) for power law generalised random graphs with exponent 2.5 and cut off 2.
(Right top) Comparison between learning curves for locally (dashed line) and globally (solid line) normalised kernels for
Erd\H{o}s-R\'enyi random graphs. (Right bottom) As (right top) for power law random graphs. \label{fig:locallc}
\label{fig:localpoisson} \label{fig:localpowerlaw} \label{fig:globallocaloverlay}}
\end{figure}

Learning curve predictions for GPs with locally normalised kernels as they result from the cavity approach described above are shown in Figure
\ref{fig:locallc}. The figure shows numerically simulated learning curves and the cavity prediction, both for Erd\H{o}s-R\'enyi random graphs (left) and  power law generalised random graphs (centre) of size $V=500$. As for the globally normalised case one sees that the cavity predictions
are quantitatively very accurate even with the simplified update Equation~\eqref{eqn:localensembleupdate}. They capture all aspects of learning curve both qualitatively and quantitatively, including, for example, the shoulder in the curves from disconnected single vertices, a feature discussed in more detail below.

\begin{figure}
  \input{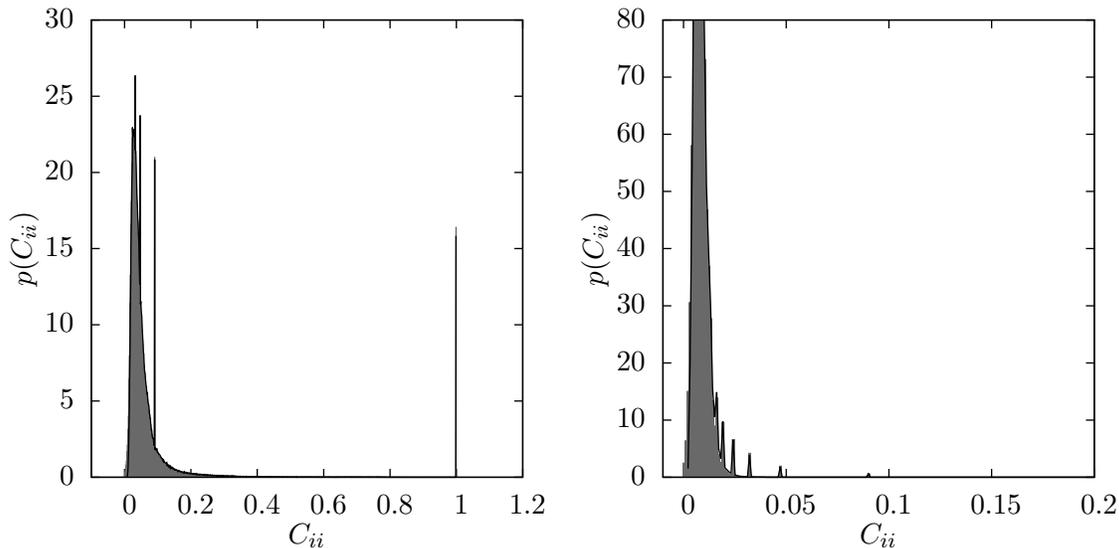}
\caption{(Left) Grey: histogram of posterior variances at $\nu=1.172$ for the locally normalised random walk kernel with $a=2$, $p=10$, averaged over ten samples each
of teacher functions, data and Erd\H{o}s-R\'enyi graphs with mean degree $\lambda=3$ and $V=1000$ vertices. Black: cavity prediction
for this distribution in the large graph limit. (Right) As (left) but for $\nu=6.210$. \label{fig:posteriorvariance}}
\end{figure}

The fact that the cavity predictions of the learning curve for a locally normalised kernel are indistinguishable from the numerically simulated learning curves in Figure \ref{fig:locallc} leads us to believe that the simplification made by dropping the consistency requirement in \eqref{eqn:localensembleupdate} is in fact exact. This is further substantiated by looking not just at the average of the posterior variance over vertices, which is the Bayes error, but its distribution across vertices. As shown in Figure \ref{fig:posteriorvariance}, the cavity predictions for this distribution are in very good agreement with the results of numerical simulations. This holds not only for the two values of $\nu$ shown, but along the entire learning curve.

\section{A Qualitative Comparison of Learning with Locally and Globally Normalised Kernels}\label{sec:qualcompare}

The cavity approach we have
developed gives very accurate predictions for learning curves for GP
regression on graphs using random walk kernels. This is true for both
global and local normalisations of the kernel. We argued in
\Sref{sec:kernelnorm} that the local normalisation is much more
plausible as a probabilistic model, because it avoids variability in
the local prior variances that is non-trivially related to the local
graph structure and so difficult to justify from prior knowledge. We
now compare what the qualitative effects of the two
different normalisations are on GP learning.

It is not a simple matter to say which kernel is
`better', the locally or globally normalised one. Since we have dealt with the matched case, where for each kernel the target functions are sampled from a GP prior with that kernel as covariance function, it would not make sense to say the better kernel is the one that gives the lower Bayes error for given number of examples, as the Bayes error reflects both the complexity of the target function and the success in learning it. A more definite answer could be obtained only empirically, by running GP regression with local and global kernel normalisation on the same data sets and comparing the prediction errors and also the marginal data likelihood. The same approach could also be tried with synthetic data sets generated from GP priors that are mismatched to both priors we have considered, defined by the globally and locally normalised kernel, though justifying what is a reasonable choice for the prior of the target function would not be easy.

While a detailed study along the lines above is outside the scope of this paper, we can nevertheless at least qualitatively study the effect of the kernel normalisation, to understand to what extent the corresponding priors define significantly different probabilistic models.
Figure \ref{fig:globallocaloverlay} (right top and bottom) overlays the learning curves for global and local kernel normalisations, for an Erd\H{o}s-R\'enyi and a power law generalised random graph respectively. There are qualitative differences in the shapes of the learning curves, with the ones for the locally normalised kernel exhibiting a shoulder around $\nu=2$. This shoulder is due to the proper normalisation of isolated vertices to unit prior variance; by contrast, as shown earlier in Figure \ref{fig:poissonvariance} (left), global normalisation gives too small a prior variance to such vertices. The inset in Figure \ref{fig:localpoisson} (left) shows the expected learning curve contributions from all locally normalised isolated vertices (single vertex subgraphs) as dotted lines. After the GP learns the rest of the graph to a sufficient accuracy, the single vertex error dominates the learning curve until these vertices have typically seen at least one example. Once this point has been passed, the dominant error comes once more from the giant connected component of the graph, and the GP learns in a similar manner to the globally normalised case. A similar effect, although not plotted, is seen for the generalised random graph case.

\begin{figure}
\input{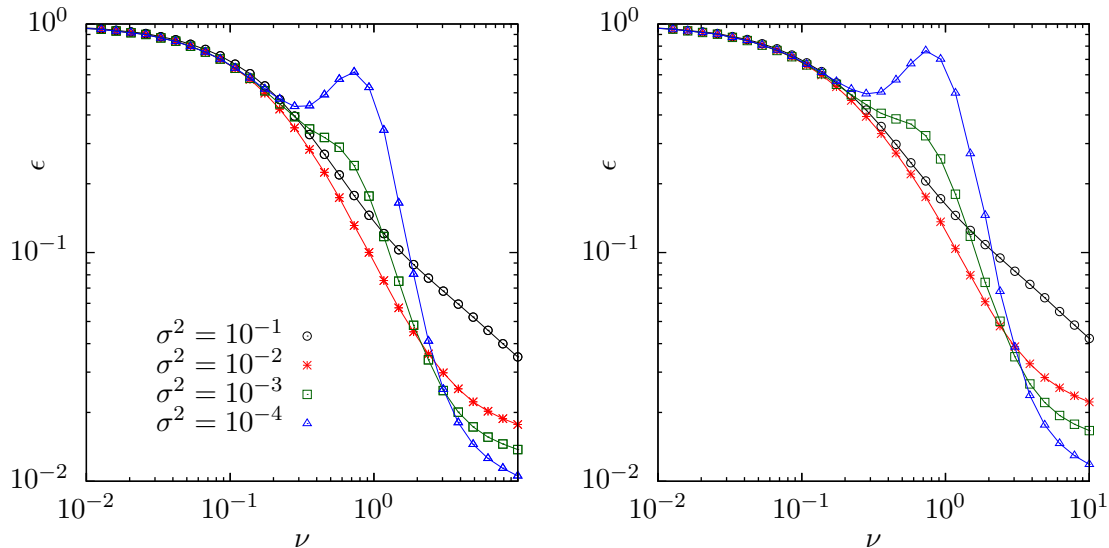}
\caption{(Left) Numerically simulated learning curves for a GP with a globally normalised kernel with $p=10$ and $a=2$, on Erd\H{o}s-R\'enyi random graphs with mean degree 3 for a range of noise levels. The teacher GP has a locally normalised kernel with the same parameters. (Right) As (left) but for power law generalised random graphs with exponent 2.5 and cutoff 2.  \label{fig:stglobalpowerlawmismatch}
\label{fig:stglobalpoissonmismatch}}
\end{figure}

We can extend the scope of this qualitative comparison by examining how a student GP with a kernel with one normalisation performs when learning from a teacher with a kernel with the other normalisation. This is a case of \emph{model mismatch}; our theory so far does not extend to this scenario, but we can obtain learning curves by numerical simulation.
Figure \ref{fig:stglobalpoissonmismatch} (left) shows the case of GP students with a globally normalised kernel learning from a teacher with a locally normalised kernel on an Erd\H{o}s-R\'enyi graph. The learning curves for the mismatched scenario are very different from those for the matched case (Figures \ref{fig:globallc} and \ref{fig:locallc}), showing an increase in error as $\nu$ approaches unity. The resulting maximum in the learning curve again emphasises that the two choices of normalisation produce distinctly different probabilistic models. Similar behaviour can be observed for the case of power law generalised random graphs, shown in Figure \ref{fig:stglobalpowerlawmismatch} (right) and for the case of GP students with a locally normalised kernel learning from a teacher with a globally normalised kernel, shown in Figure \ref{fig:stlocalpoissonmismatch}. In all cases close inspection (see Appendix \ref{app:mismatchbump}) shows that the error maximum is caused by `dangling edges' of the graph, that is, chains of vertices (with degree two) extending away from the giant graph component and terminating in a vertex of degree one.

\begin{figure}
\input{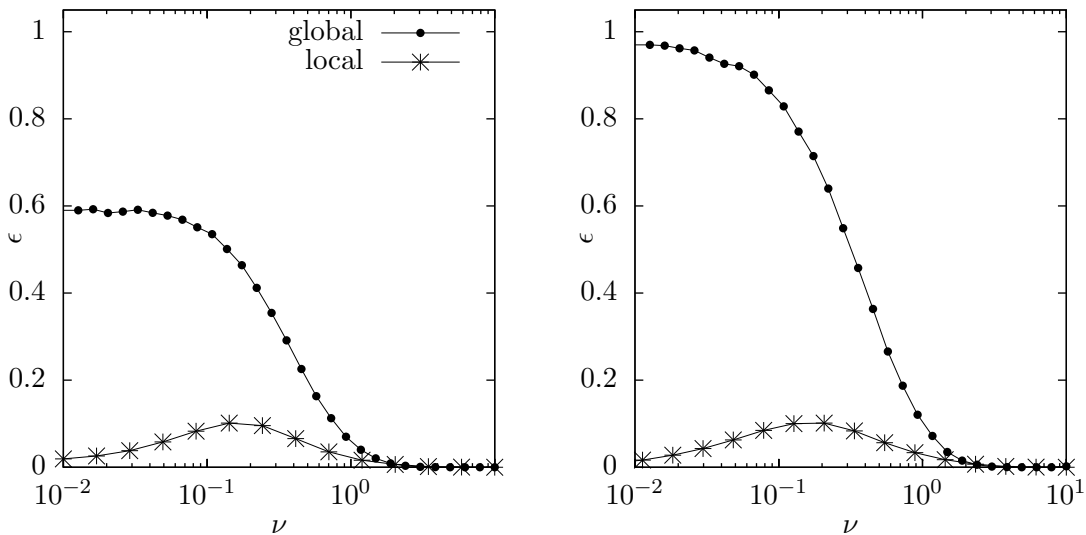}
\caption{(Left) Error variance for GP regression with locally (stars) and globally (circles) normalised kernels against $\nu$, on Erd\H{o}s-R\'enyi random graphs, and for matched learning with $p=10$, $a=2$, $\sigma^{2}=0.1$. (Right) As (left) but for power law generalised random graphs with exponent 2.5 and cutoff 2.\label{fig:poissonlocalvariances} \label{fig:powerlawlocalvariances}}
\end{figure}

As a final qualitative comparison between globally and locally normalised kernels, Figure \ref{fig:poissonlocalvariances} shows the variance of local posterior variances. This measures how much the local Bayes error typically varies from vertex-to-vertex, as a function of the data set size. Plausibly one would expect that this error variance is low initially when prediction on all vertices is equally uncertain. For large data sets the same should be true because errors on all vertices are then low. In an intermediate regime the error variance should become larger because examples will have been received on or near some vertices but not others. As Figure \ref{fig:poissonlocalvariances} shows, for kernels with local normalisation we find exactly this scenario, both for Erd\H{o}s-R\'enyi and power law random graphs. The error variance is low for small $\nu=N/V$, increasing to a peak at $\nu\approx 0.2$ and finally decreasing again.

These results can now be contrasted with those for globally normalised kernels, also displayed in Figure \ref{fig:poissonlocalvariances}. Here the error variance is largest at $\nu=0$ and decays from there. This means that the initial variance in the local prior variances is so large that any effects from the uneven distribution of example locations in any given data set remain sub-dominant throughout. We regard this as another indication of the probabilistically implausible character of the large spread of prior variances caused by global normalisation.

\begin{figure}
\input{figs/gnuplot/stlocal_mismatch.tex}
\caption{(Left) Numerically simulated learning curves for a GP with a locally normalised kernel with $p=10$ and $a=2$, on Erd\H{o}s-R\'enyi random graphs with mean degree 3 for a range of noise levels. The teacher GP has a globally normalised kernel with the same parameters. (Right) As (left) but for power law generalised random graphs with exponent 2.5 and cutoff 2.  \label{fig:stlocalpowerlawmismatch}
\label{fig:stlocalpoissonmismatch}}
\end{figure}

\section{Conclusions and Further Work}\label{sec:conclusions}

In this paper we studied random walk kernels and their application to GP regression. We began, in Section \ref{sec:randomwalk}, by studying the random walk kernel, with a focus on applying this to $d$-regular trees and graphs. We showed that the kernel exhibits a rather subtle approach to the fully correlated limit; this limit is reached only beyond a graph-size dependent threshold for the kernel range $p/a$, where cycles become important. If $p/a$ is large but below this threshold, the kernel reaches a non-trivial limiting shape.

In Section \ref{sec:lc} we moved on to the application of random walk kernels to GP regression. We showed, in Section \ref{sec:kernelnorm}, that the more typical approach to normalisation, that is, scaling the kernel globally to a desired average prior variance, results in a large spread of local prior variances that is related in a complicated manner to the graph structure; this is undesirable in a prior. We suggested as a simple remedy to perform local normalisation, where the raw kernel is normalised by its local prior variance so that the prior variance becomes the same at every vertex.

In order to get a deeper understanding of the performance of GPs with random walk kernels we then studied the learning curves, that is, the mean Bayes error as a function of the number of examples. We began in section \ref{sec:evalpred} by applying a previous approximation due to \citet{Sollich1999a} and \citet{Malzahn2005} to the case of discrete inputs that are vertices on a graph. We demonstrated numerically that this approximation is accurate only for small and large number of training examples per vertex, $\nu=N/V$, while it fails in the crossover between these two regimes. The outline derivation of this approximation suggested how one might improve it: one has to exploit fully the structure of random graphs, using the cavity method, thus avoiding approximating the average over data sets. In Section \ref{sec:cavitypred} we implemented this programme, beginning in Section \ref{sec:cavityglobal} with the case of global normalisation. We showed that by Fourier transforming the prior and introducing $2p$ additional variables at each vertex one can rewrite the partition function in terms of a complex-valued Gaussian graphical model, where the marginals that are required to calculate the Bayes error can be found using the cavity method, or equivalently belief propagation. In Section \ref{sec:cavitylocal} we tackled the more difficult scenario of a locally normalised kernel. This required two sets of cavity equations. The first serves to calculate the local normalisation factors. The second one then combines these with the information about the data set to find the local marginal distributions. One might be tempted to consider applying our methods to a lattice so that one could make an estimate of the learning curves for the continuous limit, that is, regression with a squared exponential kernel for inputs in $\mathbb{R}^2$ or similar. Sadly, however, since the cavity method requires graphs to be treelike this is not possible.

Finally in Section \ref{sec:qualcompare} we qualitatively compared GPs with kernels that are normalised globally and locally. We showed that learning curves are indeed qualitatively different. In particular, local normalisation leads to a shoulder in the learning curves owing to the correct normalisation of the single vertex disconnected graph components. We also considered numerically calculated mismatch curves. The mismatch caused a maximum to appear in the learning curves, as a result of the large differences in the teacher and student priors. Lastly we looked at the variance among the local Bayes errors, for GPs with both globally and locally normalised kernels. Plausibly, locally normalised kernels lead to this error variance being maximal for intermediate data set sizes. This reflects the variation in number of examples seen at or near individual vertices. For globally normalised kernels, the error variance inherited from the spread of local prior variances is always dominant, obscuring any signatures from the changing `coverage' of the graph by examples.

In further work we intend to extend the cavity approximation of the learning curves to the case of mismatch, where teacher and student have different kernel hyperparameters. It would also be interesting to apply the cavity method to the learning curves of GPs with random walk kernels on more general random graphs, like those considered in \citet{Rogers2010} and \citet{Kuhn2011}. This would enable us to consider graphs exhibiting some community structure. Looking further ahead, preliminary work has shown that it should be possible to extend the cavity learning curve approximation to the problem of graph mismatch, where the student has incomplete information about the graph structure of the teacher.

\appendix
\section{Random Walk Kernel on $d$-Regular Trees}\label{app:clplc}
We detail here how to derive the limiting kernel from \eqref{eqn:clpinfinity} in Section \ref{sec:dregtree} and how to calculate the scaling of the learning curve stated in Section \ref{sec:scaling}.

\subsection{Large-$p$ Limit from Heat Kernel Results}\label{app:chunglimit}
In \citet{Chung1999} the authors considered heat kernels on graphs. The heat kernel for a graph with normalised Laplacian $\bm{L}$ (see Section \ref{sec:randomwalk} in the main text) is given by
\begin{equation}
\bm{H} = \exp\left(-t\bm{L}\right),\quad t > 0,
\end{equation}
This is exactly the diffusion kernel given in \eqref{eqn:diffkernel} with $t=\frac{1}{2}\sigma^{2}$. \citet{Chung1999} gave results for heat kernels on $d$-regular trees. Their method involves treating the tree as a covering of a path, which is closely related to the mapping onto a one-dimensional lattice discussed in \Sref{sec:dregtree}. The eigenvectors and eigenvalues of $\bm{L}$ are then found within this mapping. If we call the unnormalised form of the random walk kernels we considered in the main text again $\hat{\bm{C}}= \left(\bm{I} - a^{-1}\bm{L}\right)^{p}$, then we can directly use the results of \citet{Chung1999}, simply by modifying the function that is applied to each eigenvalue $\lambda$ of $\bm{L}$ from $\exp(-t\lambda)$ to $(1-\lambda/a)^p$. This gives, for the unnormalised random walk kernel element between two vertices on the tree at distance $l$,
\begin{gather}
\hat{C}_{0,p} = \frac{2(d-1)d}{\pi}\int_{0}^{\pi}\rmd x\,\frac{[1-(1-2\cos(x)\sqrt{d-1}/d)/a]^p\sin^{2}(x)}{d^{2}-4(d-1)\cos^2(x)},\label{eqn:heat0}\\
\begin{split}
\hat{C}_{l\geq 1,p} = \frac{2}{\pi(d-1)^{l/2-1}}\int_{0}^{\pi}\rmd x\, &[1-(1-2\cos(x)\sqrt{d-1}/d)/a]^p\\&\times\frac{\sin(x)[(d-1)\sin((l+1)x) - \sin((l-1)x)]}{d^{2}-4(d-1)\cos^2(x)}\label{eqn:heatl}.
\end{split}
\end{gather}
In the limit $p\to \infty$, the factor in square brackets in both expressions means that only values of $x$ up to $O(p^{-1/2})$ contribute significantly to the integral. One can then expand $\sin(x)\approx x$ linearly everywhere, and similarly use $\cos(x)\approx 1$. Calculating $\hat{C}_{l,p}$ in this way for $p\to\infty$ gives the result \eqref{eqn:clpinfinity} for $C_{l,p}=\hat{C}_{l,p}/\hat{C}_{0,p}$.

\subsection{Scaling Form for Large $p$}\label{app:lcscale}
In \Sref{sec:scaling} we derived the learning curve decay due to the first example, given in \eqref{eqn:initial_error_decay}. As explained there, to understand how this behaves for large $p$ one needs to know how $C_{l,p}$ approaches its limiting form \eqref{eqn:clpinfinity}. In the outline derivation in the previous subsection we saw that in order to obtain this limiting form one expands not just $\sin(x)$, but also $\sin((l-1)x)$ and $\sin((l+1)x)$ linearly because $x=O(p^{-1/2})$ is small. For the two sine factors this will break down once $lp^{-1/2}$ is of order one, that is, for large enough $l$. This suggests looking at $C_{l,p}$ as function of $l'=lp^{-1/2}$. To focus on the relevant part of the integral one can similarly transform the integration variable to $x'=xp^{1/2}$. For $p\to\infty$ at constant $l'$, the integral for $\hat{C}_{l,p}$ in \eqref{eqn:heatl} then becomes proportional to
\nopagebreak
\begin{equation}
(d-1)^{-l/2}\int_{0}^{\infty}\rmd x'\, {\rm e}^{-x'{}^2\sqrt{d-1}/(da)}x'
(d-2)\sin(l'x').
\end{equation}
Integration by parts now gives $C_{l,p} \propto \hat{C}_{l,p} \propto  (d-1)^{-l/2} l' \exp(-\frac{(l')^2 d}{2\sqrt{d-1}})$. Comparing with the large-$l$ limit of \eqref{eqn:clpinfinity}, $C_{l,p}\propto (d-1)^{-l/2} l'$, shows that the effect of finite $p$ is contained in the exponential (Gaussian) cutoff factor which becomes significant for $l'$ of order unity, that is, $l = O(p^{1/2})$, as stated in the main text.

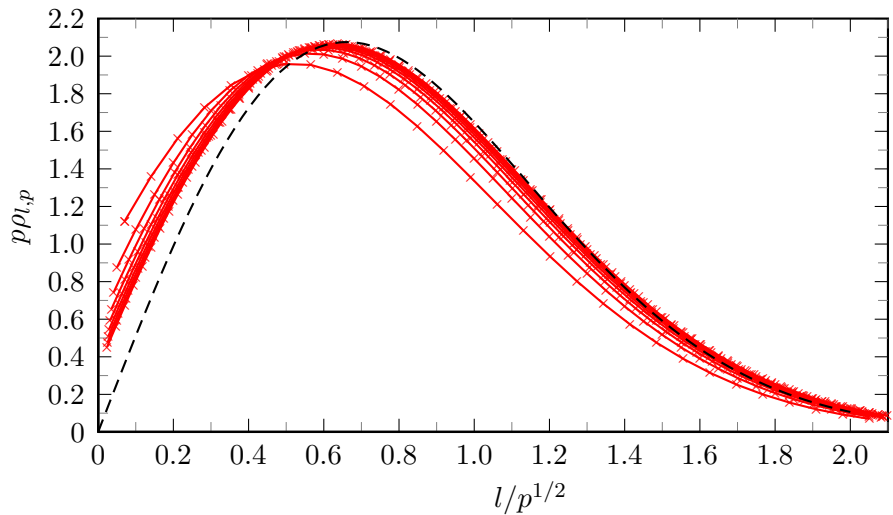
\begin{figure}
\begin{center}
\psset{yunit=2.5}
\psset{xunit=5}
\begin{pspicture}(-0.2,-0.35)(2.3,2.3)
\psaxes[Dx=0.2,Dy=0.2,labels=all,ticksize=0 5pt,subticks=2,tickstyle=inner,axesstyle=frame]{-}(0,0)(0,0)(2.1,2.2)%[$lp^{-1/2}$,0][$\rho_{l,p}p$,0]
\rput(1.15,-0.35){$l/p^{1/2}$}
\rput{90}(-0.2,1.15){$p\rho_{l,p}$}
\readdata\data{figs/gnuplot/saneappscaling0.dat}
\dataplot[linecolor=red,showpoints=true,dotstyle=x]{\data}
\readdata\data{figs/gnuplot/saneappscaling1.dat}
\dataplot[linecolor=red,showpoints=true,dotstyle=x]{\data}
\readdata\data{figs/gnuplot/saneappscaling2.dat}
\dataplot[linecolor=red,showpoints=true,dotstyle=x]{\data}
\readdata\data{figs/gnuplot/saneappscaling3.dat}
\dataplot[linecolor=red,showpoints=true,dotstyle=x]{\data}
\readdata\data{figs/gnuplot/saneappscaling4.dat}
\dataplot[linecolor=red,showpoints=true,dotstyle=x]{\data}
\readdata\data{figs/gnuplot/saneappscaling5.dat}
\dataplot[linecolor=red,showpoints=true,dotstyle=x]{\data}
\readdata\data{figs/gnuplot/saneappscaling6.dat}
\dataplot[linecolor=red,showpoints=true,dotstyle=x]{\data}
\readdata\data{figs/gnuplot/saneappscaling7.dat}
\dataplot[linecolor=red,showpoints=true,dotstyle=x]{\data}
\readdata\data{figs/gnuplot/saneappscaling8.dat}
\dataplot[linecolor=red,showpoints=true,dotstyle=x]{\data}
\psplot[linecolor=black,linestyle=dashed,plotpoints=100]{0}{2}{x -1.1445788 mul x mul EXP x mul 5.17625085 mul}
%\rput(1.2,1.3){
%\psset{yunit=0.38}
%\psset{xunit=0.38}
%\psaxes[ticks=none,labels=none]{->}(0,0)(0,0)(2.3,2.4)
%\psplot[linecolor=black,plotpoints=100]{0}{2}{x 0 0.7 GAUSS x mul 9 mul}
%}
\end{pspicture}
\end{center}
\caption{Plot demonstrating the $p\to\infty$ scaling from of $p\rho_{l,p}$, the normalised version of $\hat{R}_{l,p}$. Dashed line shows the analytically found functional from of the scaling.\label{fig:appscaling}}
\end{figure}

Visually, the cutoff effects for large $l$ and $p$ can be seen more clearly by plotting not $C_{l,p}$ directly but rather $\hat{R}_{l,p}$, the unormalised version of $R_{l,p}$ as defined in Section \ref{sec:scaling}. As $v_l=d(d-1)^{l-1}$ for $l\geq 1$, the additional $\sqrt{v_l}$ factor just removes the decay with $(d-1)^{-l/2}$ from $C_{l,p}$. From the above discussion, we then expect to find for large $l$ that $\hat{R}_{l,p} \propto l' \exp(-\frac{(l')^2 d}{2\sqrt{d-1}})$.
A plot of numerical results for a suitably normalised version of $\hat{R}_{l,p}$ (see below), plotted against $lp^{-1/2}$ for increasing $p$, is shown in Figure \ref{fig:appscaling}. There is a clear trend towards the predicted asymptotic behaviour for large $p$ (dashed line).

It may be somewhat surprising that the cutoff lengthscale that appears in the analysis above is $l=O(p^{1/2})$, which for large $p$ is much smaller than the typical kernel range $p/a$. To understand this intuitively, one can go back to \eqref{eqn:Rshelljump} for $R_{l,p}$ and use that this is essentially unnormalised diffusion. Taking $\hat{R}_{l,p}$ as the unnormalised version of $R_{l,p}$ again and letting $\rho_{l,p} = \hat{R}_{l,p}\gamma^{-p}$ with $\gamma = (1-1/a) +
2\sqrt{d-1}/(ad)$ to re-establish normalisation, one finds that $\rho_{l,p}$ evolves almost according to a diffusion process in `time' $p$, except that probability conservation is broken at $l=0$ and $l=1$. In the (leaky) diffusion process $\rho_{l,p}$ the typical lengthscale $l$ should then scale with time as $p^{1/2}$, exactly as we found above. This diffusion interpretation can also be used to reproduce the quantitative scaling form, by adapting the methods of \citet{Monthus1996} where the authors study a one dimensional walk with a reflective boundary to compute the large-$p$ scaling. The normalised version of $\hat{R}_{l,p}$ shown in Figure \ref{fig:appscaling} is in fact $p\rho_{l,p}$.

\section{Analysis of the Mismatch Learning Curves}\label{app:mismatchbump}
In this appendix we suggest a way of understanding the mismatch maximum seen in the learning curves plotted in Figure \ref{fig:stglobalpoissonmismatch} (left) and (right) and Figure \ref{fig:stlocalpoissonmismatch} (left) and (right). We begin by considering the mean of the GP posterior, that is, the prediction function. If we arrange the predictive means at each vertex into a vector $\bar{\f}$, then from \eqref{eqn:GPmean} this can be written as
\begin{equation}\label{eqn:fft}
\bar{\f} = \K_{V}\K^{-1}\y,
\end{equation}
where $(K_{V})_{j\nu} = C_{j,x_\nu}$ for $\nu=1,\ldots,N$ and $j=1,\ldots,V$ and $K_{\mu\nu}=C_{x_\mu,x_\nu} + \sigma^2\delta_{\mu,\nu}$ for $\mu,\nu=1,\ldots,N$. (Recall that $x_\mu$ is the location of the $\mu$-th training input, which on a graph is just a label in the range $\{1,\ldots,V\}$.) We can rewrite this in the form $\bar{\f}=\M\z$ where
\begin{gather}
\M = \K_{V}\K^{-1/2},\\
\z = \K^{-1/2}\y.
\end{gather}
One sees that $\M$ represents teacher-independent aspects of $\bar{\f}$.
The vector of `pseudo-training outputs' $\z$ has been defined so that in the {\em matched} case, it obeys $\langle\z\z\T\rangle = \I$. We think of $\z$ as pseudo-training outputs because if its components $z_\mu$ are sampled as independent and identically distributed unit variance Gaussian random variables, then $\bar{\f}=\M\z$ will have the correct distribution of mean prediction vectors. The columns $\m_1,\ldots,\m_N$ of $\M=\begin{pmatrix}\m_{1}&\cdots&\m_N\end{pmatrix}$ represent the `effective prediction vectors' conjugate to the $z_\mu$. In the matched case each $\bar{f}_i^2$ will then be, on average, a sum of squares of the corresponding entries in the effective prediction vectors.

\begin{figure}
\begin{center}
\readdata\match{figs/gnuplot/stglobal_zmatchsq_x_col.dat}
\readdata\mismatch{figs/gnuplot/stglobal_zmismatchsq_x_col.dat}
\psset{llx=-1.4cm}
\psset{xAxisLabel={example},xAxisLabelPos={100,-1.2},yAxisLabel={$\z\z\T$},yAxisLabelPos={-50,-2.5},ticksize=0 5pt,subticks=2,tickstyle=inner,axesstyle=frame}
\begin{psgraph}[Dx=50,Dy=1,Ox=0,Oy=-8,ylogBase=10](0,-9)(0,-9)(213,3){5.0cm}{7.6cm}
\listplot[linecolor=red,plotstyle=LineToXAxis,linewidth=0.5pt]{\mismatch}
\listplot[plotstyle=LineToXAxis,linewidth=0.5pt]{\match}
\psdot[linecolor=darkgreen,dotsize=4pt](182,2.603)
\end{psgraph}
\includegraphics{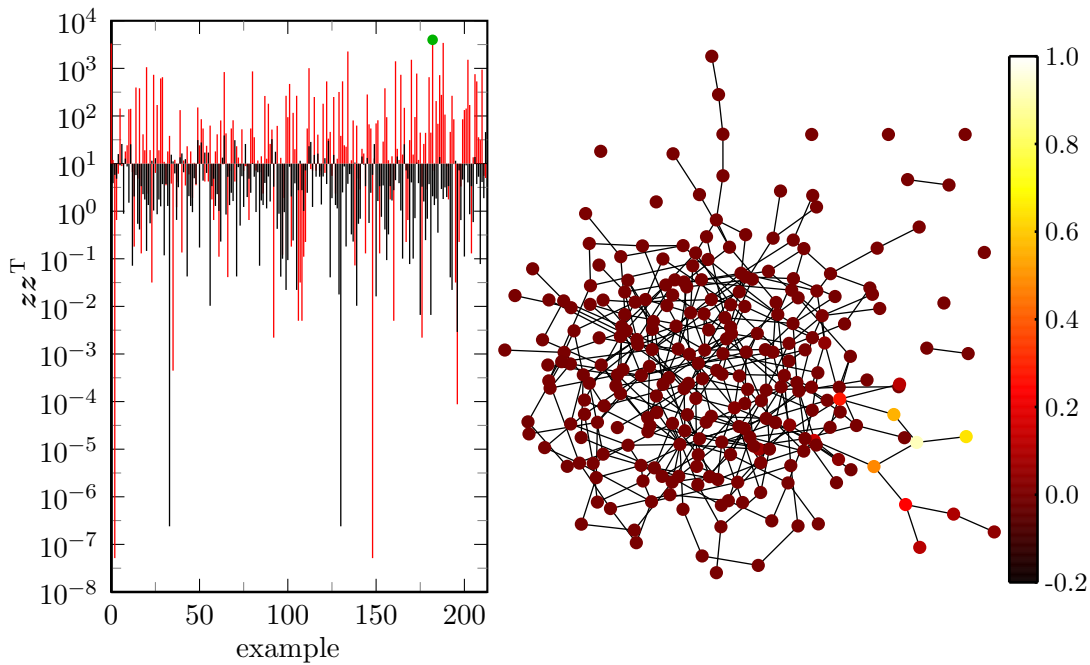}
\vspace{0.5cm}
\end{center}
\caption{(Left) Black lines show diagonal elements of $\z\z\T$ for a typical draw from the Erd\H{o}s-R\'enyi random graph ensemble with average degree 3 in a matched scenario with 250 vertices, $p=10$, $a=2$ and $\sigma^2=0.0001$. Red lines show $\z\z\T$ for the case where student is globally normalised and the teacher is locally normalised. (Right) A plot of the `prediction vector' for the largest spike  in $\z\z\T$ (indicated by a green circle in the left hand plot) under mismatch.\label{fig:stglobalzzT}}
\end{figure}

\begin{figure}
\begin{center}
\readdata\match{figs/gnuplot/stlocal_zmatchsq_x_col.dat}
\readdata\mismatch{figs/gnuplot/stlocal_zmismatchsq_x_col.dat}
\psset{llx=-1.4cm}
\psset{xAxisLabel={example},xAxisLabelPos={100,-1.2},yAxisLabel={$\z\z\T$},yAxisLabelPos={-50,-1.5},ticksize=0 5pt,subticks=2,tickstyle=inner,axesstyle=frame}
\begin{psgraph}[Dx=50,Dy=1,Ox=0,Oy=-8,ylogBase=10](0,-8)(0,-8)(213,4){5.0cm}{7.6cm}
\listplot[linecolor=red,plotstyle=LineToXAxis,linewidth=0.5pt]{\mismatch}
%\listplot[plotstyle=LineToXAxis,linewidth=0.5pt]{\match}
\psdot[linecolor=darkgreen,dotsize=4pt](206,3.454)
\end{psgraph}
\includegraphics{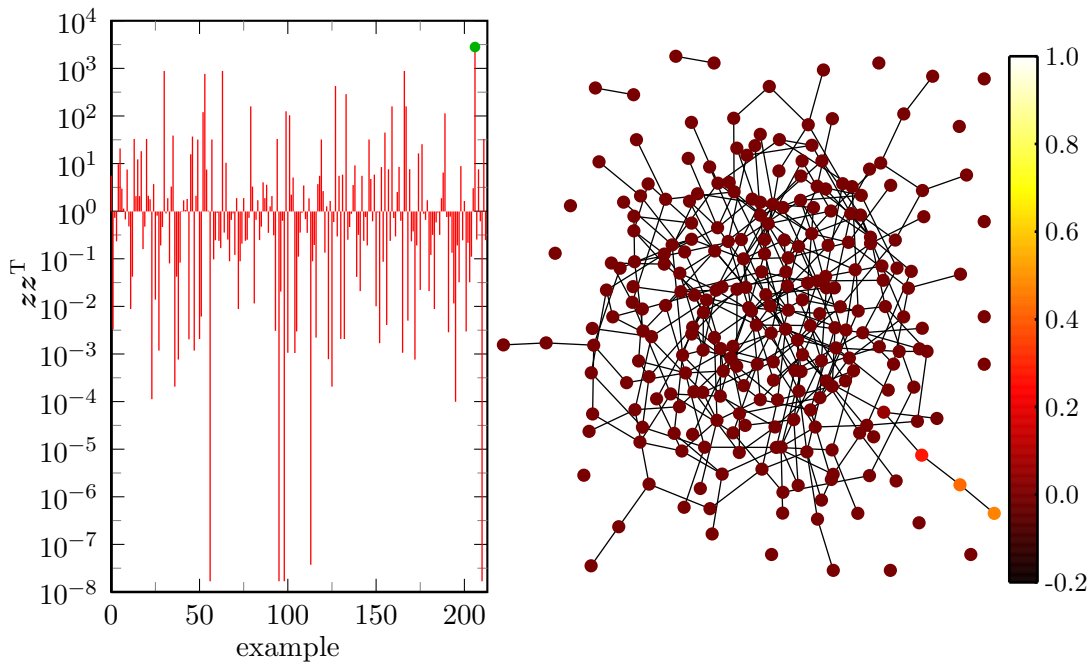}
\vspace{0.5cm}
\end{center}
\caption{Analogue of Figure \protect\ref{fig:stglobalzzT} for a locally normalised student and globally normalised teacher\label{fig:stlocalzzT}}
\end{figure}

For the case of mismatch, for example, because student and teacher use differently normalised kernels, the only change is in the statistics of $\z$. Their covariance matrix $\langle \z\z\T\rangle$ will no longer simply be the identity matrix, and so act as a re-weighting of the prediction vectors. To understand our numerical simulations, we considered a value of $\nu$ around the maximum in the mismatched learning curve, listed the largest (diagonal) entries of $\z\z\T$ and plotted their corresponding prediction vectors. These prediction vectors were generally localised around dangling ends of the graph, substantiating our claim in the main text that it is from these graph regions that the major contribution to the learning curve maximum arises. Typical plots of $\z\z^T$ and the prediction vector corresponding to the largest spike in $\z\z\T$ for an instance of an Erd\H{o}s-R\'enyi graph with 250 vertices are shown in Figures \ref{fig:stglobalzzT} and \ref{fig:stlocalzzT} for both a locally normalised teacher and globally normalised student and a globally normalised teacher and locally normalised student respectively. Both these plots clearly show the largest spike corresponding to a prediction vector localised on a dangling edge in the graph. Similar plots of the other large spikes give prediction vectors localised on other dangling edges.

\bibliography{refs.bib}
\end{document}